\documentclass[10pt,twocolumn,letterpaper]{article}

\usepackage{cvpr}      
\definecolor{cvprblue}{rgb}{0.21,0.49,0.74}
\usepackage[pagebackref,breaklinks,colorlinks,allcolors=cvprblue]{hyperref}

\usepackage{multirow}
\usepackage{color}
\usepackage{amsmath}
\usepackage{bm}
\usepackage[table]{xcolor}

\usepackage{graphicx}
\usepackage{makecell}
\usepackage{placeins}
\usepackage{stfloats}
\usepackage[
  enable,
  labname={CMLab},
  logo={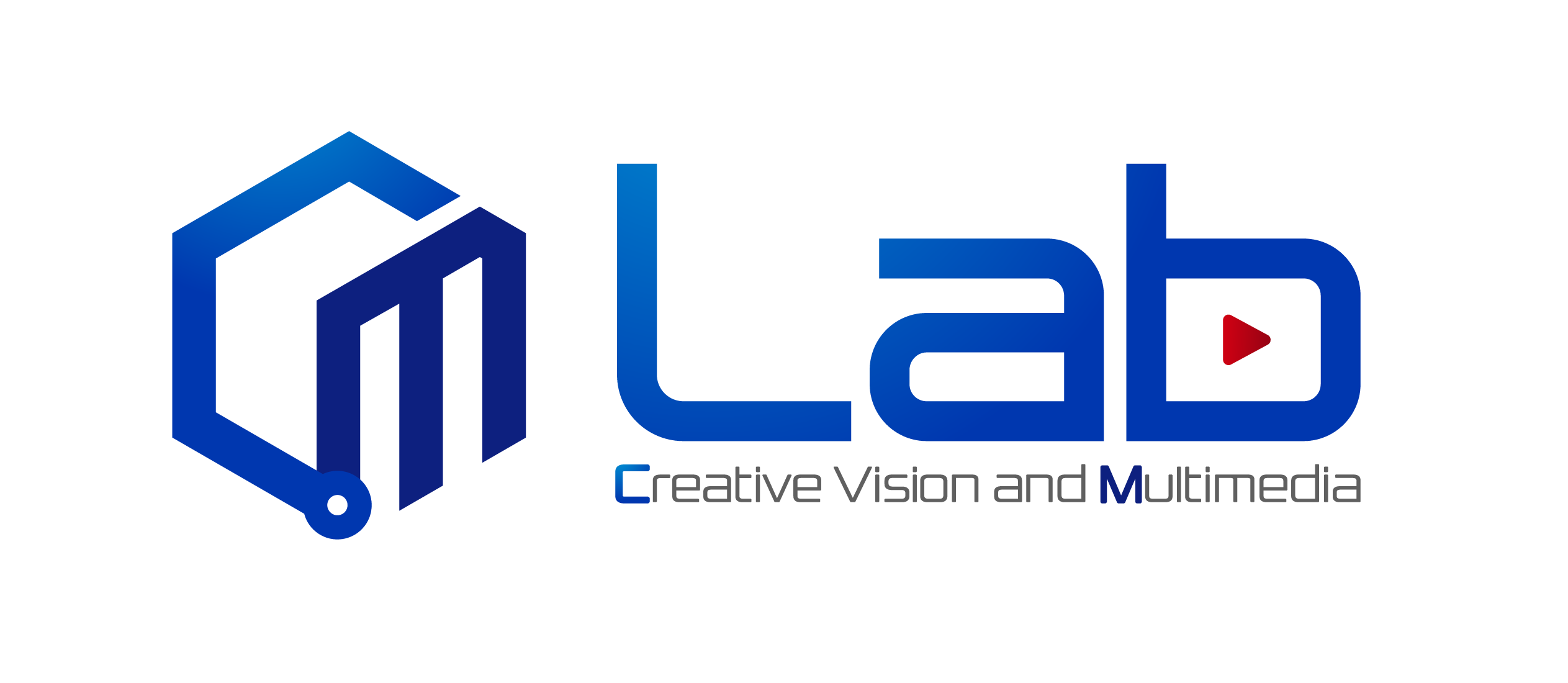},
  titleicon={assets/icon.png},
  logowidth={22mm},
  projectpage={https://cmlab-korea.github.io/FLAIR/}
]{cmlab}

\definecolor{red}{rgb}{1,0.2,0.2}
\definecolor{or}{rgb}{1,0.5,0.25}
\definecolor{green}{rgb}{0, 1, 0}
\definecolor{bl}{rgb}{0, 0, 1}
\definecolor{brown}{rgb}{0.59, 0.3, 0}
\definecolor{cyan}{rgb}{0, 1, 1}
\definecolor{c_lowbest}{rgb}{1.0,1.0,0.9}
\definecolor{c_highbest}{rgb}{0.9,1.0,0.9}

\newcommand{\best}[1]  {\textcolor{red}{\textbf{#1}}}
\newcommand{\second}[1]  {\textcolor{blue}{\underline{#1}}}

\def\confName{CVPR}
\def\confYear{2026}

\title{FLAIR: Frequency- and Locality-Aware Implicit Neural Representations}

\cmlabAuthors{
Sukhun Ko$^{1}$ \qquad 
Seokhyun Youn$^{1}$ \qquad 
Dahyeon Kye$^{1}$ \qquad 
Kyle Min$^{2}$ \qquad 
Chanho Eom$^{3}$ \qquad 
Jihyong Oh$^{\dagger,1}$
}

\cmlabAffiliations{
$^{1}$ CMLab, Chung-Ang University \qquad 
$^{2}$ Oracle \qquad 
$^{3}$ Perceptual AI LAB, Chung-Ang University
}

\cmlabAuthorEmail{
\{looloo330, hisn16, rpekgus, jihyongoh\}@cau.ac.kr \qquad 
kylemin@umich.edu \qquad 
cheom@cau.ac.kr
}

\cmlabProjectPage{https://cmlab-korea.github.io/FLAIR/}

\begin{document}
\maketitle
\let\thefootnote\relax
\footnotetext{$^{\dagger}$Corresponding author}
\begingroup
\makeatletter
\@ifundefined{thefootnote}{}{\renewcommand{\thefootnote}{}}
\makeatother
\endgroup

\vspace{-1.5em}
\vspace{-3em}
\begin{abstract}
Implicit Neural Representations (INRs) leverage neural networks to map coordinates to corresponding signals, enabling continuous and compact representations. This paradigm has driven significant advances in various vision tasks. However, existing INRs lack frequency selectivity and spatial localization, leading to an over-reliance on redundant signal components. Consequently, they exhibit spectral bias, tending to learn low-frequency components early while struggling to capture fine high-frequency details.
To address these issues, we propose FLAIR (Frequency- and Locality-Aware Implicit Neural Representations), which incorporates two key innovations. The first is Band-Localized Activation (BLA), a novel activation designed for joint frequency selection and spatial localization under the constraints of the time-frequency uncertainty principle (TFUP). Through structured frequency control and spatially localized responses, BLA effectively mitigates spectral bias and enhances training stability. The second is Wavelet-Energy-Guided Encoding (WEGE), which leverages the discrete wavelet transform to compute energy scores and explicitly guide frequency information to the network, enabling precise frequency selection and adaptive band control. Our method consistently outperforms existing INRs in 2D image representation, as well as 3D shape reconstruction and novel view synthesis.
\end{abstract}
\vspace{-0.5cm}
\section{Introduction}
\label{sec:intro}

\begin{figure}[t]
    \centering
    \includegraphics[width=\linewidth,keepaspectratio]{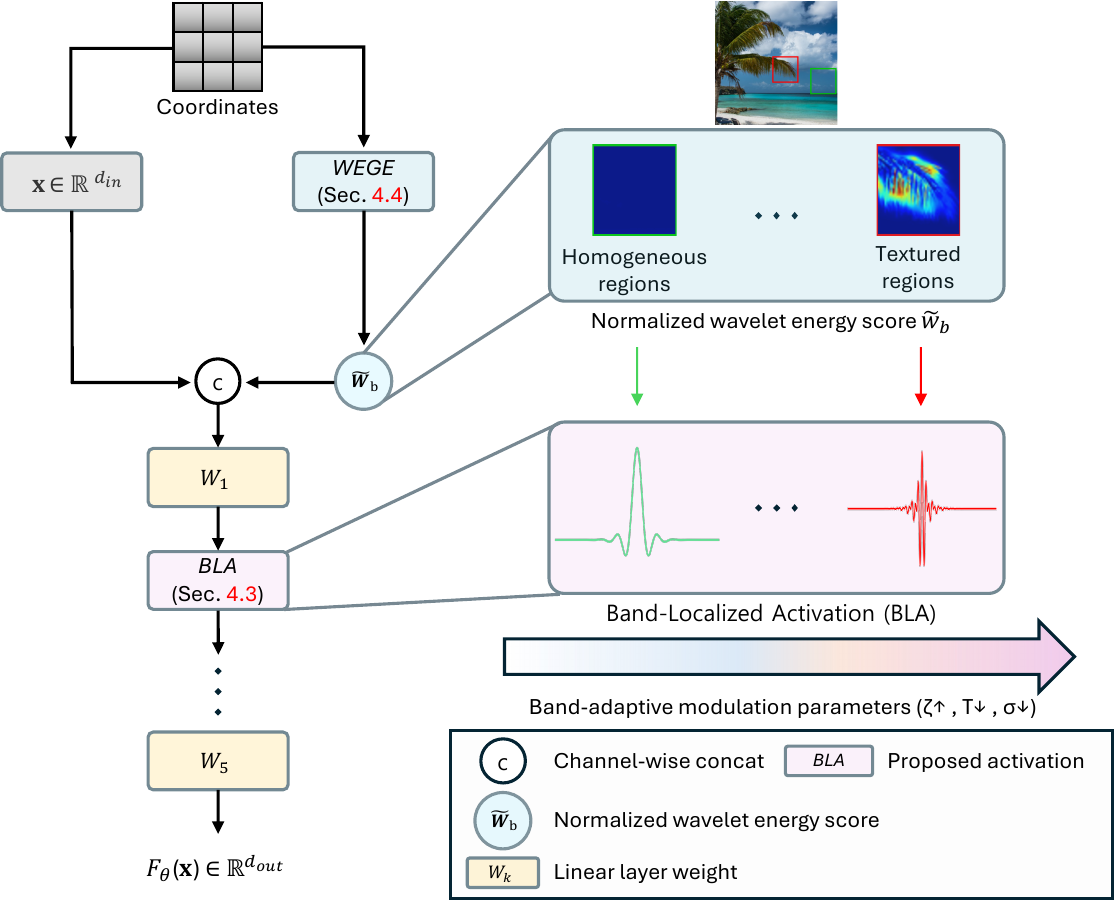}
    \caption{
    \textbf{Overall architecture of FLAIR.} FLAIR consists of two complementary components: Band-Localized Activation (BLA) and a Wavelet-Energy-Guided Encoding (WEGE) module. 
    \textbf{Right:} WEGE computes normalized wavelet-energy scores $\tilde{w}_b$ over the input coordinates, highlighting spatial frequency characteristics by assigning lower wavelet-energy scores to homogeneous regions (green box) and higher scores to textured regions (red box).
    \textbf{Left:} Wavelet-energy scores $\tilde{w}_b$ are channel-wise concatenated with input coordinates and passed through Band-Localized Activation (BLA). BLA modulates the signal representation via learnable band-adaptive parameters ($\zeta, T, \sigma$), enabling frequency shifting and band-limiting across low- and high-frequency components.
    } 
    \label{fig:teaser}
    \vspace{-0.5cm}
\end{figure}

Traditional explicit representations, including discrete grid-based methods~\cite{choy20163d,liu2020neural,guo2020deep} such as volumes and unstructured formats like point clouds, have contributed to solving vision problems~\cite{tewari2022advances,ulyanov2018deep}. However, these approaches have increasingly shown limitations in addressing the diverse and ill-posed inverse problems~\cite{demoment2002image,yan2017image}.
To overcome these limitations, implicit neural representations (INRs)~\cite{sitzmann2020implicit} model signals as continuous mappings from input coordinates to their corresponding values via neural networks, thereby supporting continuous attribute queries and facilitating the seamless integration of differentiable physical processes.
This coordinate-based paradigm has enabled notable advances in previously challenging tasks such as super-resolution (SR)~\cite{chen2021learning} and denoising~\cite{buades2005review}. More recently, it has been extended to higher-dimensional problems including 3D shape reconstruction~\cite{park2019deepsdf,mescheder2019occupancy} and neural radiance fields~\cite{mildenhall2021nerf}.

However, existing INR methods commonly suffer from the spectral bias problem~\cite{rahaman2019spectral}, where networks tend to learn low-frequency components first. This leads to an inability to recover fine details, a challenge that remains unresolved. To mitigate the spectral bias, various positional encoding (PE) schemes have been proposed, such as Sinusoidal PE~\cite{tancik2020fourier} and Wavelet PE~\cite{zhao2025adaptive}, which embed input coordinates into a higher-dimensional space using a fixed set of basis functions.
Although these approaches embed coordinate information into a fixed set of $M$ bases, their representational capacity is fundamentally limited by the number of bases $M$, as well as by their learnable range. These limitations constrain their expressive capacity and often lead to imperfect representations~\cite{rahimi2007random}.

Beyond predefined frequency bases, recent studies~\cite{sitzmann2020implicit,serrano2024hosc} have attempted to mitigate spectral bias by enhancing the expressiveness of activation functions. However, these methods have primarily focused on broadening the domain of activation functions~\cite{liu2024finer,rezaeian2025sl2a}, rather than enabling precise frequency selection. As a result, networks may construct overlapping bases, lacking the ability to selectively target the necessary frequency bands. This limitation impairs their effectiveness in real-world inverse problems, such as denoising~\cite{yan2024boosting}, where the suppression of irrelevant frequency bands (\textit{e.g.}, noise) is essential for accurate signal recovery.

To address these limitations, we propose \textbf{FLAIR} (Frequency- and Locality-Aware Implicit Neural Representations), as illustrated in Fig.~\ref{fig:teaser}. FLAIR integrates two complementary components: (i) Band-Localized Activation (BLA) for joint frequency-domain selection and time-domain localization, and (ii) Wavelet-Energy-Guided Encoding (WEGE) for region-adaptive frequency guidance. 

Specifically, we adopt a band-limited formulation that enables explicit frequency selection for signal modeling. 
However, increasing the degree of band limitation inevitably induces oscillations in the time domain, leading to training instability and degraded time-domain localization~\cite{gottlieb1997gibbs}.
To mitigate these issues, instead of directly employing the ideal band-limited function such as \texttt{sinc}~\cite{saratchandran2024sampling}, we newly propose the Band-Localized Activation (BLA). Our design is based on the band-limiting term for frequency control and on transition-smoothing and localization terms, which are responsible for improving training stability and adaptively governing spatial localization, as illustrated in Fig.~\ref{fig:BLA} (c). 
Although the time–frequency uncertainty principle (TFUP) states that perfect localization in both domains is unattainable, our formulation allows the model to adaptively learn the optimal trade-off between frequency selectivity and spatial localization via learnable parameters.

As the second key component of FLAIR, we propose Wavelet-Energy-Guided Encoding (WEGE), which adaptively provides explicit information to quantify continuous-frequency components, indicating whether a region is high- or low-frequency. This allows the model to perform region-adaptive frequency selection, thereby enhancing its representational flexibility. Notably, WEGE requires only 0.1K additional parameters and is compatible with plug-and-play integration into other activation-based architectures. In particular, it complements BLA by providing explicit frequency information that enhances its frequency selection capability. In summary, our contributions are as follows: 
\begin{itemize}
 \setlength\itemsep{0.1cm}
  \item We propose FLAIR, a novel INR architecture that integrates two complementary components: (i) Band-Localized Activation (BLA) and (ii) Wavelet-Energy-Guided Encoding (WEGE).

  \item We introduce BLA, a novel activation function that mitigates spectral bias through frequency selection, while suppressing oscillations and enhancing time localization under TFUP.
  
  \item We introduce WEGE, which provides quantified continuous-frequency information, enabling BLA to perform region-adaptive frequency selection. Importantly, WEGE requires only 0.1K additional parameters and is compatible with plug-and-play integration.

  \item  We achieve state-of-the-art performance across diverse INR tasks,
  including 2D image fitting and restoration, 3D signed distance field,
  and 5D neural radiance field.



\end{itemize}

\section{Related Work}
\label{sec:related_work}

\begin{figure*}
\centering
\includegraphics[width=\linewidth,keepaspectratio]{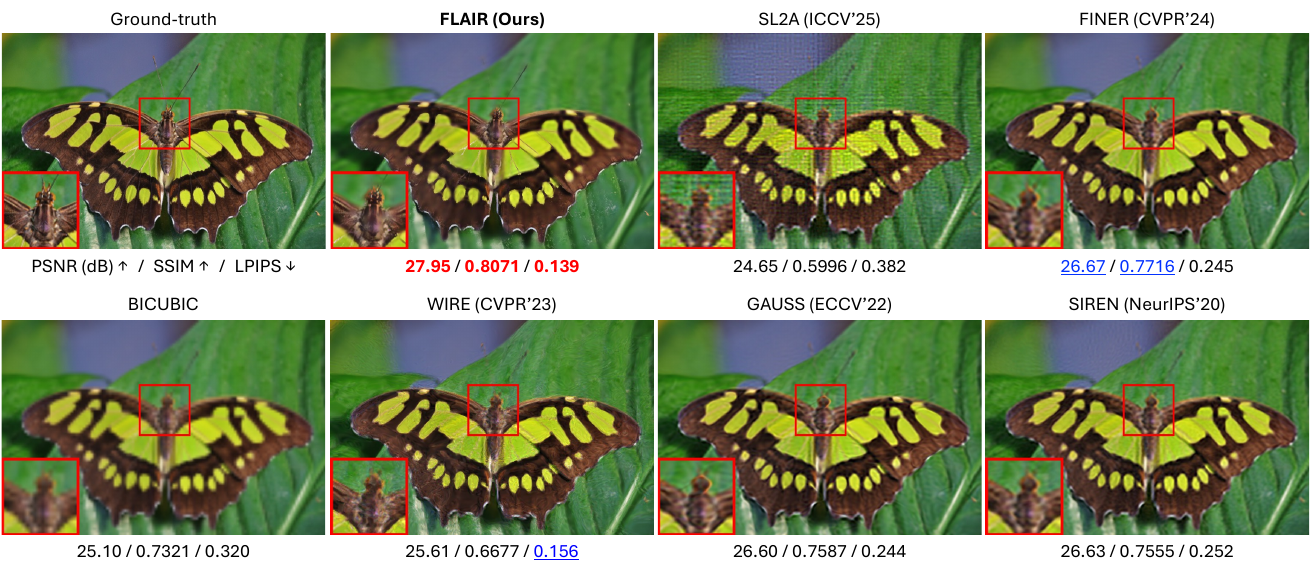}
\vspace{-0.7cm}
\caption{\textbf{Qualitative comparisons of ×4 super-resolution.} 
\best{Red} and \second{blue} denote the best and second-best performances, respectively.}
\label{fig:SR}
\vspace{-0.3cm}
\end{figure*}

\noindent \textbf{Implicit Neural Representations.} Standard MLPs with ReLU activations~\cite{agarap2018deep} exhibit a strong bias towards learning low-frequency content due to the non-periodic nature of ReLU, limiting their ability to capture fine details, a phenomenon known as spectral bias~\cite{rahaman2019spectral}. To address this spectral bias~\cite{cai2024batch,shi2024improved,shiinductive}, a series of periodic activation functions has been proposed, most notably SIREN~\cite{sitzmann2020implicit} and FINER~\cite{liu2024finer}. SIREN employs sinusoidal activations to model high-frequency components, but its performance is highly dependent on the choice of hyperparameters and is sensitive to initialization, necessitating careful design choices to avoid suboptimal results. FINER introduces a variable-periodic sine activation and tunes the supported frequency set by adjusting the initialization range of the bias vector. However, in practice, both SIREN and FINER remain limited in addressing spectral bias due to their reliance on fixed global harmonic bases, which restrict flexible spectral adaptation during training.

Alternatively, wavelet-based activations such as WIRE~\cite{saragadam2023wire} have aimed to address spectral bias through spatial compactness~\cite{chng2022gaussian}, offering improved interpretability and spatial control. Nonetheless, these approaches remain limited in their ability to fully overcome spectral bias, as they do not offer explicit frequency selection.

More recently, filter-based activation functions have been actively investigated in the context of INRs. A recent theoretical analysis~\cite{saratchandran2024sampling} has suggested that \texttt{sinc} functions, as a type of filter-based activation, can serve as an optimal choice for INR tasks.
Despite their theoretical advantages, these filter-based functions exhibit inherent limitations. In the spatial domain, their infinite oscillations can induce training instability and visual artifacts~\cite{jerri1977shannon,gottlieb1997gibbs}, and in the absence of frequency-shift mechanisms, they only act as low-pass~\cite{stenger1981numerical}, resulting in limited ability to generalize across diverse vision tasks~\cite{park2019deepsdf,chen2021learning,mildenhall2021nerf}.

To address these challenges, we propose BLA (Sec.~\ref{sec:BLA}), a key component of FLAIR that enables joint spatial localization and frequency selection. By operating under the TFUP, BLA allows the model to adaptively balance spatial localization and frequency selectivity during training, thereby mitigating spectral bias and overcoming the aforementioned limitations.

\noindent \textbf{Frequency-Guided Representations and Conditioning.} Frequency-guided representations and conditioning methods have been actively explored to enhance the modeling of task-relevant frequency components, which are essential for preserving fine details in low-level vision tasks~\cite{li2023multi, wang2024frequency}. For example, Local Texture Estimator for Implicit Representation Function (LTE)~\cite{lee2022local} employs dedicated modules composed of convolutional neural networks (CNN)~\cite{krizhevsky2012imagenet} and fully connected (FC) layers~\cite{basha2020impact} to separately estimate key frequency components, such as amplitude, frequency, and phase, allowing the network to infer dominant frequencies and the corresponding Fourier coefficients.

In parallel, the Local Implicit Wavelet Transformer (LIWT)~\cite{duan2024local} leverages DWT to decompose input features into low- and high-frequency components, which are further refined to enhance frequency-aware representation.
More recently, Zhao \etal~\cite{zhao2025adaptive} introduce a method that adaptively places local wavelet bases at high-frequency regions, enabling efficient encoding of fine details.

Building on this line of research, we propose a novel encoding, WEGE (Sec.~\ref{sec:WEGE}), specifically designed to work with activation functions in INRs. Unlike methods~\cite{lee2022local,duan2024local} that introduce additional auxiliary networks and increase optimization complexity, which hinders fast convergence required for INR settings, our approach leverages a non-parametric DWT-based prior to guide region-adaptive frequency modulation, with the entire WEGE module introducing only 0.1K parameters.

\vspace{-1.0em}
\section{Preliminary: Time-Frequency Uncertainty Principle (TFUP)}
\label{sec:preliminary}
The mathematical foundation of the time-frequency uncertainty principle (TFUP) in signal processing originates from the Heisenberg uncertainty principle~\cite{busch2007heisenberg} in quantum mechanics, which states that there is a fundamental limit to simultaneously and precisely measuring a particle’s position and momentum given by $\Delta x\,\Delta p \ge \hbar/2$, where $\Delta x$ and $\Delta p$ are the standard deviations of position and momentum, and $\hbar=h/(2\pi)$ is the reduced Planck constant.

Extending this concept to the domain of signal processing~\cite{parhizkar2015sequences}, the TFUP states that for any signal $x(t)$, the product of its temporal standard deviation $\sigma_t$ and frequency standard deviation $\sigma_f$ is bounded below by $\sigma_t \sigma_f \ge \frac{1}{4\pi}$, where $\sigma_t$ and $\sigma_f$ quantify the temporal and frequency uncertainty of $x(t)$, respectively. This relationship implies that a signal highly localized in time (small $\sigma_t$) must necessarily have a broad frequency spectrum (large $\sigma_f$), and vice versa. Thus, the uncertainty principle imposes a fundamental limit on the simultaneous localization of a signal in both time and frequency domains.

\section{Proposed Method: FLAIR}
\label{sec:proposed_method}

In this section, we first provide a brief formulation of an Implicit Neural Representations (INRs). Grounded in the Riesz basis framework, \texttt{sinc} functions are optimal band-limited bases, yet their global support introduces instability, motivating localized designs for INRs. Building on these foundations, we present a novel framework, FLAIR, composed of two key components: (i) Band-Localized Activation (BLA)~(Sec.~\ref{sec:BLA}), a band-limited activation for frequency-domain selection and time-domain localization, and (ii) Wavelet-Energy-Guided Encoding (WEGE)~(Sec.~\ref{sec:WEGE}) for region-adaptive frequency guidance.

\begin{figure*}
\centering
\includegraphics[width=\linewidth,keepaspectratio]{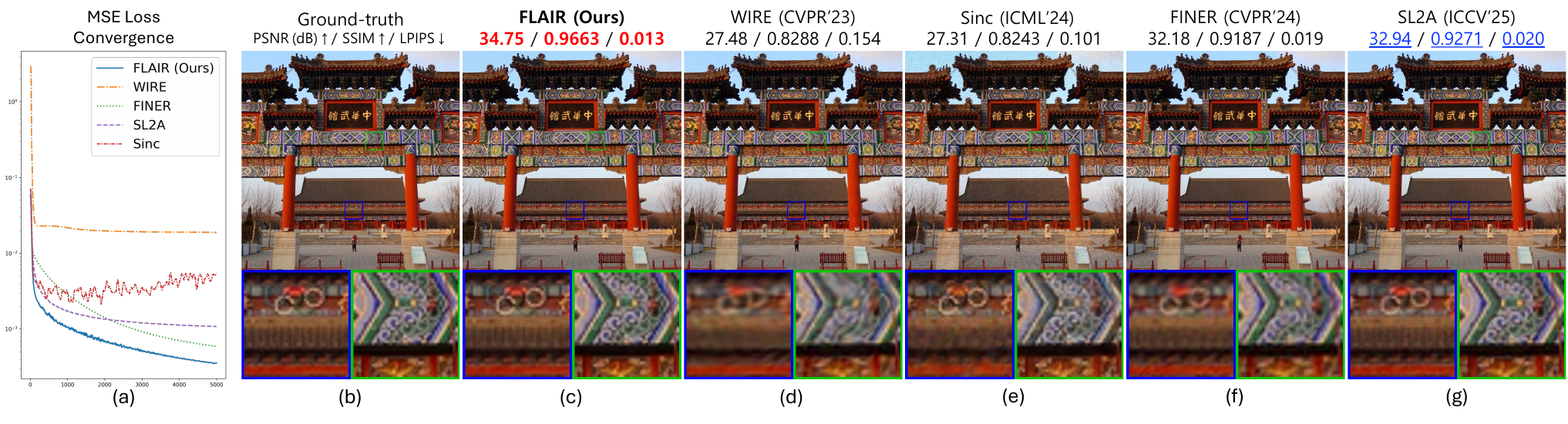}
\vspace{-0.8cm}
\caption{\textbf{(a) MSE loss convergence. (b) Ground-truth. (c)–(g) Comparison between FLAIR and other methods on image fitting.}}
\vspace{-0.2cm}
\label{fig:fitting}
\end{figure*}

\subsection{Formulation of an INR}
\label{sec:INR}
An Implicit Neural Representation (INR) aims to represent a continuous signal by directly mapping input coordinates to their corresponding signal values through a neural network $F_{\theta}$. 
Formally, the function is defined as $F_{\theta} : \mathbb{R}^{d_{\text{in}}} \rightarrow \mathbb{R}^{d_{\text{out}}},\ \mathbf{x} \mapsto F_{\theta}(\mathbf{x})$, where $\theta \in \mathbb{R}^M$ denotes all learnable parameters of the network. 
The input $\mathbf{x} \in \mathbb{R}^{d_{\text{in}}}$ corresponds to a spatial coordinate,  such as $(x, y)$ in 2D or $(x, y, z)$ in 3D.  
The output $F_{\theta}(\mathbf{x}) \in \mathbb{R}^{d_{\text{out}}}$ represents the signal value at that location (e.g., RGB color or volumetric density $\sigma$).
The training data is given as a set of pairs $\mathcal{D} = \{ (\mathbf{x}_i, \mathbf{y}_i) \}_{i=1}^N$, where $\mathbf{y}_i$ denotes the ground-truth signal at coordinate $\mathbf{x}_i$. 
The network is optimized by minimizing a standard mean squared error (MSE) loss:
\begin{equation}
\mathcal{L}(\theta) = \frac{1}{N} \sum_{i=1}^N \|F_{\theta}(\mathbf{x}_i) - \mathbf{y}_i\|_2^2,
\label{eq:inr_loss}
\end{equation}
\noindent 
where \(N\) denotes the number of training samples.
This simple yet general formulation enables INRs to model complex signals with high fidelity. 
However, accurate reconstruction of visual signals requires stronger control over both frequency selection and spatial localization. To address this, we optimize $F_{\theta}$ by introducing the Band-Localized Activation (BLA), which achieves these properties.

\subsection{Instability of Sinc Activations}
\label{sec:Instability}

\begin{figure*}
\centering
\includegraphics[width=0.90\linewidth]{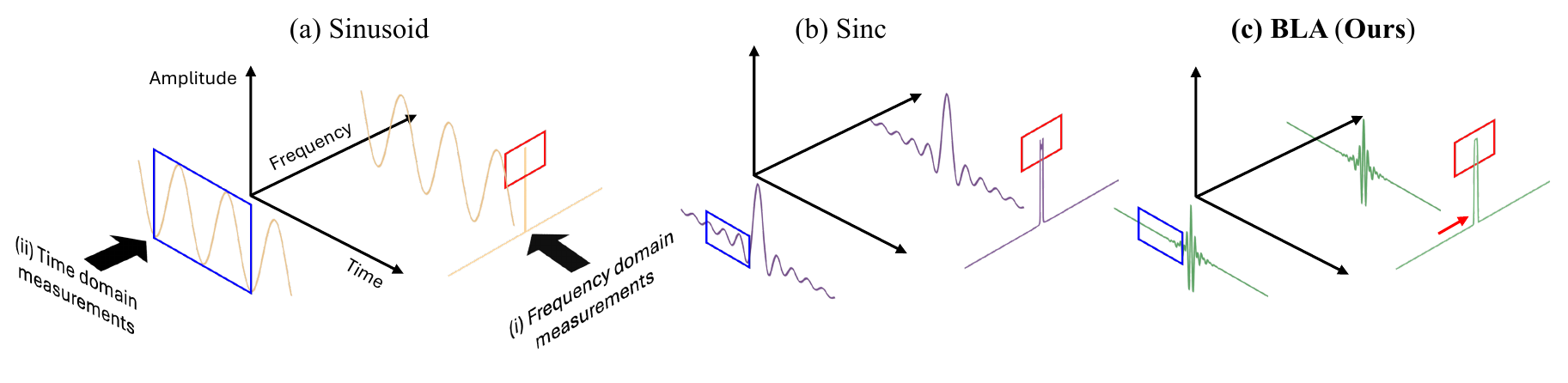}
\vspace{-0.7cm}
\caption{\textbf{Both precise frequency selection and time localization under the TFUP.}
In domain (i), Sinusoid (a) produces only two fixed frequency components, so representing composite signals with diverse spectra requires many hidden dimensions. \texttt{Sinc} (b) provides sharp band selectivity. Our BLA (c) retains similar selectivity to (b) and further introduces a learnable frequency shift parameter $\zeta$ to access higher frequency bands (\textcolor{red}{red box}). 
In domain (ii), while (a) and (b) exhibit global support and oscillations (\textcolor{blue}{blue box}), BLA (c) yields localized responses and reduces oscillations, mitigating noise propagation. Finally, our BLA (c) jointly modulates the frequency–time trade-off under the TFUP through its learnable parameters $(\zeta, T, \sigma)$, achieving both precise frequency selectivity and time localization.}
\vspace{-0.3cm}
\label{fig:BLA}
\end{figure*}

Building on Riesz basis framework~\cite{christensen2003introduction,unser2002sampling},
Saratchandran \textit{et al}.~\cite{saratchandran2024sampling} show that any 
signal $s \in L^{2}(\mathbb{R})$ can be approximated within an error bound $\epsilon$, 
and further establish that the \texttt{sinc} function is theoretically optimal.
 
Despite this theoretical optimality, such guarantees do not translate into stable 
optimization. The instability appears in the gradient of the loss $\mathcal{L}$ with respect to the weights $W_{ij}$, driven by the residual and the oscillatory term induced by the derivative of the sinc activation:
{\small
\begin{equation}
\begin{aligned}
\frac{\partial \mathcal{L}}{\partial W_{ij}}
= & -\frac{2}{\pi (W_i x)^2}
\int 
\underbrace{[g(x) - f(x;W)]}_{\textit{residual}} 
\cdot x_j
\\[0.2em]
&\cdot
\underbrace{
\left[
\frac{\pi W_i x}{T}\cos\left(\frac{\pi W_i x}{T}\right)
- \sin\left(\frac{\pi W_i x}{T}\right)
\right]
}_{\textit{oscillatory}}
\, dx .
\end{aligned}
\label{eq:oscillation}
\end{equation}
}
\noindent
Here, $g(x)$ is the target function and $f(x;W)$ is the network output.
Intuitively, the residual term inevitably arises during training, and the derivative of the sinc function (oscillatory term), having global support, propagates the residual error across the entire gradient field. This interaction amplifies instability throughout optimization, as shown in Fig.~\ref{fig:fitting} (a).

\subsection{Proposed Band-Localized Activation (BLA)}
\label{sec:BLA}

To mitigate the aforementioned oscillatory instability, 
we propose the Band-Localized Activation (BLA), which suppresses the global oscillation 
behavior in Eq.~\eqref{eq:oscillation} by imposing the exponential bound 
$\exp\!\left(-\frac{(W_i x)^2}{T^2\sigma^2}\right)$.
Conceptually, BLA attenuates residual error propagation through localized gradient responses, resulting in improved stability. Full derivations are provided in the ~\textit{Supplementary}.

Beyond the gradient-level analysis, the time–frequency domain offers an intuitive view of the underlying behavior. As shown in Fig.~\ref{fig:BLA} (c) red box, BLA exploits the sharp cut-off behavior for frequency selectivity while suppressing infinite oscillations (blue box), thereby enabling improved time localization. Formally, BLA is expressed as a modulated basis function:
\begin{equation}
\psi^{\mathrm{BLA}}(x)
= \phi^{\mathrm{basis}}(x)
\cdot 
\underbrace{\exp\!\left(2\pi \zeta x j\right)}_{\textit{frequency-shifting}},
\label{eq:BLA}
\end{equation}

\noindent
where $\zeta \in \mathbb{R}$ denotes the modulation frequency, and $j$ is the imaginary unit. Note that all the above functions are defined in the time domain. Importantly, multiplying by $\exp(2\pi j \zeta x)$ in the time domain is equivalent to shifting the frequency support of the basis function to $\zeta$ in the Fourier domain. This allows the basis to be centered at any desired frequency, enabling the network to adaptively capture both low- and high-frequency components as needed. In practice, $\zeta$ is treated as a learnable parameter.
The basis function $\phi^{\mathrm{basis}}$ is defined as:

{
\vspace{-0.6em}
\begin{equation}
\scalebox{0.9}{$
\phi^{\mathrm{basis}}(x) =
\smash[b]{\underbrace{\frac{\texttt{sinc}\!\left(\frac{x}{T}\right)}{T}}_{\text{\textit{band-limiting}}}}
\cdot
\smash[b]{\underbrace{\frac{\cos\!\left(\frac{\pi\beta x}{T}\right)}
     {1 - \left(\frac{2\beta x}{T}\right)^2}}_{\text{\textit{transition-smoothing}}}}
\cdot
\smash[b]{\underbrace{\exp\!\left(-\frac{x^2}{2\sigma^2}\right)}_{\text{\textit{localization}}}}.
$}
\label{eq:basis}
\end{equation}
\vspace{+1.3em}  
}

\noindent
Here, $T>0$ is the bandlimit parameter that controls the overall bandwidth, and $\beta$ is the roll-off factor in the transition-smoothing term, fixed at $0.05$ to preserve the inherent sharpness of the band-limited cutoff. 
Finally, $\sigma>0$ is the scale parameter in the localization term, adaptively governing spatial localization. We treat $T$, $\zeta$, and $\sigma$ as learnable parameters while keeping $\beta$ fixed, and the effect of each parameter on frequency and spatial localization is illustrated in Fig.~\ref{fig:scene}. 
For our BLA, we initialize \(T{=}1.0\), \(\zeta{=}1.0\), and \(\sigma{=}2.0\), with further initialization details provided in the \textit{Supplementary}.

We employ this function as the activation in each layer of our implicit neural representation (INR) network:
\begin{equation}
\begin{aligned}
\vec{z}^{\,0} &= \vec{x}, \\
\vec{z}^{\,l} &= \psi \big( W^l \vec{z}^{\,l-1} + \vec{b}^{\,l} \big), 
\quad l = 1, 2, \ldots, L-1, \\
f(\vec{x}; \theta) &= W^L \vec{z}^{\,L-1} + \vec{b}^{\,L}.
\end{aligned}
\end{equation}
\noindent
$\vec{x}$ denotes the input coordinate to the network and
$\psi(\cdot)$ is our proposed activation function, BLA.
$\vec{z}^{\,l}$ indicates the output of layer $l$. $L$ denotes the total number of layers in the network, and $\theta = \{W^l, b^l \mid l = 1, \dots, L\}$ denotes the set of learnable parameters. 
This structure enables each layer to adaptively exploit both frequency selectivity and spatial localization, 
resulting in a more expressive INR.
\begin{figure} \centering \includegraphics[width=\linewidth,keepaspectratio]{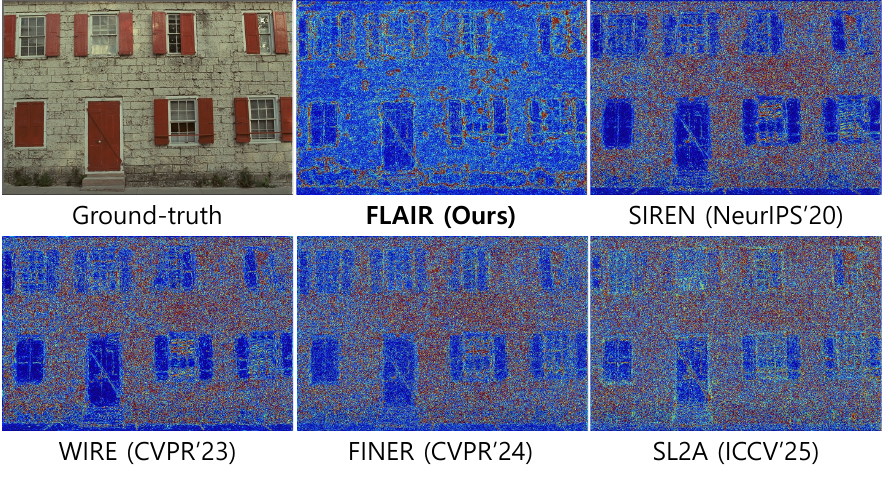}
\vspace{-0.8cm}
\caption{\textbf{Residual error heatmaps on Kodak~01.} Per-pixel absolute errors (RGB-averaged) are normalized to [0, 1]. Blue indicates lower error, and red indicates higher error. FLAIR consistently achieves lower reconstruction error than other methods.}
\vspace{-0.5cm}
\label{fig:Error} 
\end{figure}
\subsection{Wavelet-Energy-Guided Encoding (WEGE)}
\label{sec:WEGE}
To reduce artifacts in smooth regions from frequency leakage and to alleviate spectral bias by explicitly learning only the target frequency bands in detailed areas, we introduce Wavelet-Energy-Guided Encoding (WEGE). In INR settings where fast convergence is required, WEGE avoids additional auxiliary networks that increase optimization complexity, instead utilizing non-parametric discrete wavelet transform (DWT) and guided filtering~\cite{he2012guided}. Specifically, WEGE adaptively provides explicit information that quantifies continuous-frequency components, effectively indicating the relative dominance of high- or low-frequency content in each region. Our method first computes a pixel-wise energy score $R_b$ for each spatial coordinate $(x, y)$ of the input image by applying DWT. Unlike the Fourier transform, DWT jointly captures spatial and spectral information, enabling position-aware scoring. Formally, we decompose the input image $I_k$ into high- and low-frequency components via DWT and compute $R_b$ as:
\begin{equation}
\begin{aligned}
D_k^H, \; D_k^L & = \mathrm{DWT}(I_k), \\
R_b            & = \mathrm{IDWT}(D_k^H) - \mathrm{IDWT}(D_k^L).
\end{aligned}
\label{eq:dwt_rb}
\end{equation}
\noindent
The components of $D_k^H$ correspond to the high-frequency sub-bands $\{HL,\,LH,\,HH\}$, while $D_k^L$ corresponds to the low-frequency sub-band $\{LL\}$.
The residual map $R_b$ emphasizes spatial structures such as edges by subtracting the low-frequency components. To obtain a pixel-wise energy score map $E_b$, we compute it by averaging the squared $R_b$ across channels. We then normalize $E_b$ using its $E_{\min}$ and $E_{\max}$ values over the image, yielding a normalized energy score $w_b$ that preserves edge-aware structures.

Although the $w_b$-based normalized energy score map exhibits edge-aware properties, pixel-wise scoring often results in discontinuities, consequently producing noisy artifacts in the RGB output domain (details in \textit{Supplementary}). To address this, we apply a filtering operation to the normalized scores in order to suppress such score discontinuities.

Accordingly, the guided filtering~\cite{he2012guided} operation $\mathcal{G}$ is applied to the normalized energy score map $w_b$ to mitigate discontinuities while preserving edges, formulated as:
\begin{equation}
\tilde{w}_b(x) = \mathcal{G}(I(x), w_b(x); r, \epsilon),
\label{eq:dwt_whb}
\end{equation}
where $I(x)$ is the gray-scale guidance image, $w_b(x)$ is the normalized wavelet energy score at pixel $x$, $r$ is the window radius, and $\epsilon$ is the regularization parameter. The effects of these hyperparameters ($r$ and $\epsilon$) are evaluated in Table~\ref{tab:wege}.
The resulting filtered energy score map $\tilde{w}_b$, obtained by applying the guided filtering in Eq.~\eqref{eq:dwt_whb}, is channel concatenated with the original spatial coordinates $(x, y)$ and fed as input to the INR network $S(x, y) = f_\theta(x, y, \tilde{w}_b)$, where $f_{\theta}$ denotes the INR network parameterized by $\theta$ and internally employs the proposed activation BLA to adaptively select frequencies based on the local characteristics inferred from WEGE. 
This frequency-aware modulation enables capturing both high-frequency details and low-frequency structures of the target signal as shown in Fig.~\ref{fig:Error}.

\begin{figure*}[t]
    \centering
    \includegraphics[width=\linewidth,keepaspectratio]{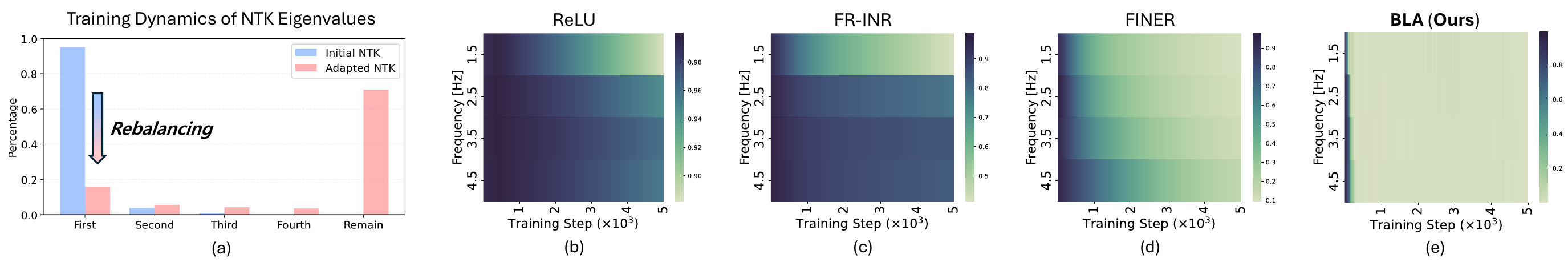}
    \vspace{-0.6cm}
    \caption{
    \textbf{Eigenvalue distribution of NTKs (a) and frequency-specific error analysis (b)–(e).}
    In (a), the training dynamics of BLA are illustrated via its empirical NTK. The x-axis enumerates the ranked eigenvalues, with “First” denoting the percentage contribution of the largest eigenvalue, and the y-axis indicates the percentage assigned to each component. In (b)-(e), we visualize the frequency-specific approximation error, where the x-axis corresponds to the training step and the y-axis to the frequency (Hz). (b)-(d) reduce low-frequency error early while learning higher-frequency components slowly, whereas our BLA (e) captures multiple frequency bands early.
    }
    \vspace{-0.2cm}
    \label{fig:ntk}
\end{figure*}

\subsection{Neural Tangent Kernel Perspective}

Neural Tangent Kernel (NTK) theory~\cite{jacot2018neural} provides a principled framework to interpret neural network training as kernel regression.
The convergence dynamics are fundamentally governed by the distribution of NTK eigenvalues, 
where larger eigenvalues correspond to components that are learned more rapidly and thus fitted more accurately~\cite{bai2023physics}.

However, the assumptions of standard NTK theory do not directly extend to practical training regimes~\cite{jacot2018neural}.
We therefore adopt the empirical NTK formulation commonly employed in recent studies~\cite{ortiz2021can,yuce2022structured,shi2024improved}. 
For samples \(x_i\) and \(x_j\), the empirical NTK is defined as
$
k'_{\mathrm{NTK}}(x_i, x_j)
=
J_{f_\Theta}(x_i)\, J_{f_\Theta}(x_j)^{\mathsf T}.$
Here, \(J_{f_\Theta}(x_i)\) denotes the Jacobian matrix of the network output \(f_\Theta\) with respect to its parameters at the \(i\)-th sample \(x_i\), 
and \(k'_{\mathrm{NTK}}(x_i, x_j)\) corresponds to the \((i,j)\)-entry of the resulting empirical NTK matrix.

We use this empirical NTK to analyze how BLA 
influences the distribution of its NTK eigenvalues during training.
Following~\cite{shi2024improved,rezaeian2025sl2a}, we use the 1D target function $f(x)=2R\!\left(\frac{\sin(3\pi x)+\sin(5\pi x)+\sin(7\pi x)+\sin(9\pi x)}{2}\right)$, where $R(\cdot)$ is a rounding function that increases the approximation complexity. Fig.~\ref{fig:ntk} (a) shows the resulting NTK eigenvalue distributions of our proposed BLA. 

At initialization, the blue bars reveal a strong concentration on the “First’’ eigenmode due to the band-limiting term in Eq.~\eqref{eq:basis}, capturing over 90\% of the spectral mass. During training, however, the learnable frequency-shifting term in Eq.~\eqref{eq:BLA} gradually moves the frequency band toward the target components. As a result, BLA redistributes the spectral mass across the “Remain’’ eigenmodes (red).

The broadened eigenvalue distribution observed in Fig.~\ref{fig:ntk} (a) suggests that BLA may mitigate the spectral bias by distributing its capacity more evenly across frequencies. To verify this, we further examine the frequency-domain behavior of BLA. In Fig.~\ref{fig:ntk} (b)–(e), we follow the observation in prior work~\cite{xu2018understanding} that MLPs exhibit a spectral bias, where low-frequency errors decrease much faster than high-frequency ones. We quantify this effect using the relative discrepancy $\Delta_k = \frac{|\mathcal{F}_D[g](k) - \mathcal{F}_D[f_\Theta](k)|}{|\mathcal{F}_D[g](k)|}$.
As shown in Fig.~\ref{fig:ntk} (e), our BLA achieves consistently low approximation errors across the entire frequency spectrum in contrast to other activations. This behavior aligns with the NTK analysis in (a), where BLA’s eigenvalue redistribution mitigates the spectral bias, as reflected in (e).

\vspace{-0.5cm}
\section{Experiments}
\label{sec:experiments}
\noindent \textbf{Implementation Details.} We thoroughly explore all possible combinations of learning rates ($1\times10^{-2}$, $5\times10^{-3}$, $1\times10^{-3}$, $5\times10^{-4}$, $1\times10^{-4}$) for each of the state-of-the-art (SOTA) models 
to choose each best configuration for fair comparisons. For our model, we use a learning rate of $5\times10^{-4}$ in all experiments. For activation functions that are sensitive to initialization, such as SIREN (where the initial $\omega_0$ is set to a large value, \textit{e.g.}, 30), we follow each method's recommended setting. All models are trained with 4 hidden layers and 256 hidden features per layer, and the number of training iterations is kept consistent across all models. All experiments are conducted on an RTX A6000 GPU. To ensure fair comparison, results are averaged over 5 random seeds. The variance is consistently small across activations, so standard deviations are omitted for brevity.
\begin{figure*}
\centering
\includegraphics[width=\linewidth,keepaspectratio]{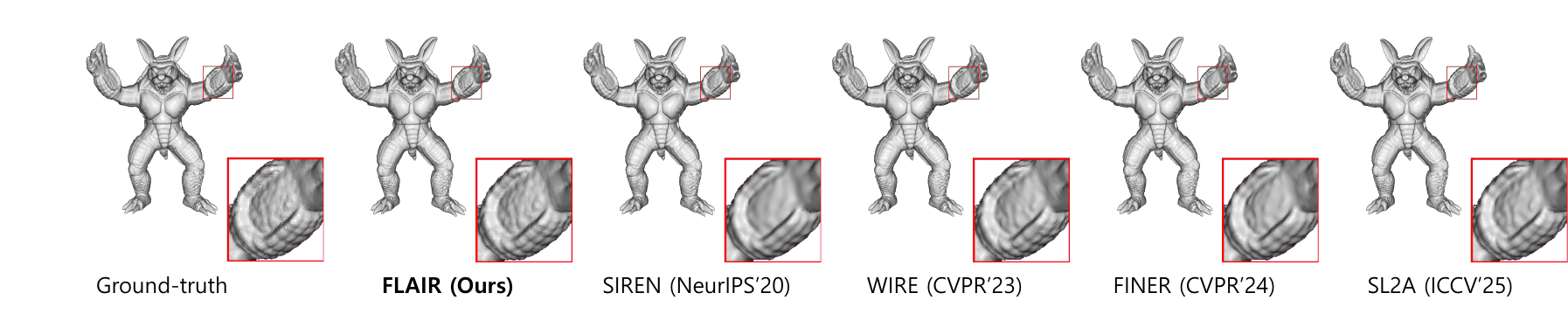}
\vspace{-0.7cm}
\caption{\textbf{Qualitative comparisons on representing the signed distance field of Armadillo.}}
\vspace{-0.3cm}
\label{fig:SDF}
\end{figure*}
\begin{figure*}
\centering
\includegraphics[width=\linewidth,keepaspectratio]{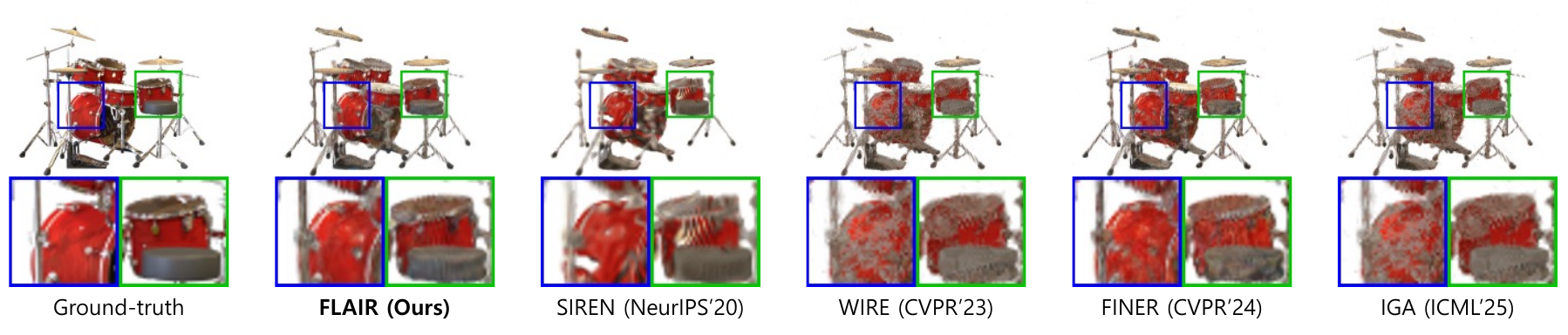}
\vspace{-0.6cm}
\caption{\textbf{Qualitative comparisons between FLAIR and the baselines on NeRF.} We follow WIRE~\cite{saragadam2023wire}, downsampling images to 200×200 and training the radiance field using only 25 input views instead of the default 100 images.}
\vspace{-0.2cm}
\label{fig:NeRF}
\end{figure*}

\begin{table}[t]
\centering
\vspace{-0.2cm}
\caption{\textbf{Quantitative image-fitting comparisons on Kodak (24 images) and DIV2K (16 images).}
\best{Red} and \second{blue} indicate best and second-best per column.}
\vspace{-0.3cm}
\label{tab:fitting}
\renewcommand{\arraystretch}{1.2}
\setlength{\tabcolsep}{4.5pt}
\resizebox{\linewidth}{!}{%
\begin{tabular}{l|c|ccc|ccc}
\hline
\multirow{2}{*}{Methods} & \multirow{2}{*}{\#Params (K)$\downarrow$} &
\multicolumn{3}{c|}{Kodak (24 images)} & \multicolumn{3}{c}{DIV2K (16 images)} \\
 &  & PSNR$\uparrow$ & SSIM$\uparrow$ & LPIPS$\downarrow$ & PSNR$\uparrow$ & SSIM$\uparrow$ & LPIPS$\downarrow$ \\
\hline
SL2A (ICCV’25)   & 330.2 & \second{36.14} & 0.9304 & 0.060 & \second{36.26} & 0.9481 & 0.034 \\
IGA (ICML’25)    & 205.1 & 34.49 & \second{0.9596} & \second{0.056} & 33.51 & \second{0.9742} & \second{0.022} \\
FR (CVPR’24)     & 6299.9 & 27.62 & 0.8781 & 0.151 & 26.22 & 0.9001 & 0.124 \\
FINER (CVPR’24)  & 198.9 & 35.46 & 0.9246 & 0.065 & 35.01 & 0.9457 & 0.039 \\
WIRE (CVPR’23)   & 91.6  & 28.80 & 0.7928 & 0.333 & 32.17 & 0.8874 & 0.107 \\
SIREN (NeurIPS’20) & 198.9 & 28.64 & 0.7792 & 0.369 & 34.36 & 0.9388 & 0.050 \\
\textbf{FLAIR (Ours)} & 199.0 & \best{37.12} & \best{0.9644} & \best{0.054} & \best{38.19} & \best{0.9754} & \best{0.016} \\
\hline
\end{tabular}}
\vspace{-0.3cm}
\end{table}

\begin{table}[t]
\centering
\vspace{-0.2cm}
\caption{\textbf{Quantitative comparisons on representing signed distance fields.} 
We report Chamfer Distance $\downarrow$ and IoU $\uparrow$.}
\vspace{-0.25cm}
\label{tab:sdf}
\renewcommand{\arraystretch}{1.2}
\setlength\tabcolsep{2.5pt}

\resizebox{\linewidth}{!}{%
\begin{tabular}{clcccccc}
\toprule
 & Methods & Armadillo & Dragon & Lucy & Thai Statue & BeardedMan & Avg. \\
\midrule
\multirow{5}*{\rotatebox{90}{Chamfer $\downarrow$}}
& SL2A (ICCV’25)
  & \second{3.235e-6} & 2.987e-5 & \second{1.972e-6} & \second{2.988e-6} & \second{3.615e-6} & 8.336e-6 \\
& FINER (CVPR’24)
  & 3.368e-6 & 2.846e-5 & 2.526e-6 & 3.931e-6 & 3.838e-6 & 8.425e-6 \\
& WIRE (CVPR’23)
  & 3.278e-6 & 2.884e-5 & 2.039e-6 & 3.143e-6 & 3.778e-6 & 8.216e-6 \\
& SIREN (NeurIPS’20)
  & 3.838e-6 & \second{2.698e-5} & 2.540e-6 & 3.358e-6 & 4.338e-6 & \second{8.212e-6} \\
& \textbf{FLAIR (Ours)}
  & \best{3.207e-6} & \best{2.656e-5} & \best{1.906e-6} & \best{2.972e-6} & \best{3.547e-6} & \best{7.637e-6} \\
\midrule
\multirow{5}*{\rotatebox{90}{IoU $\uparrow$}}
& SL2A (ICCV’25)        & \second{9.899e-1} & \second{9.849e-1} & \best{9.810e-1} & 9.657e-1 & \second{9.925e-1} & \second{9.828e-1} \\
& FINER (CVPR’24)       & 9.897e-1 & 9.749e-1 & 9.701e-1 & 9.545e-1 & 9.902e-1 & 9.759e-1 \\
& WIRE (CVPR’23)        & 9.893e-1 & 9.821e-1 & 9.778e-1 & \second{9.687e-1} & 9.908e-1 & 9.817e-1 \\
& SIREN (NeurIPS’20)    & 9.828e-1 & 9.709e-1 & 9.657e-1 & 9.669e-1 & 9.888e-1 & 9.750e-1 \\
& \textbf{FLAIR (Ours)} & \best{9.903e-1} & \best{9.851e-1} & \second{9.802e-1} & \best{9.688e-1} & \best{9.945e-1} & \best{9.838e-1} \\
\bottomrule
\end{tabular}}
\vspace{-0.4cm}
\end{table}

\subsection{Signal Representation}
\label{sec:signal}
\noindent \textbf{2D Image Representation.}
For the 2D image representation task, we use the widely adopted Kodak~\cite{kodak1999} dataset and DIV2K~\cite{agustsson2017ntire}, which are widely recognized benchmarks for image representation. Results after 5,000 training iterations are summarized in Table~\ref{tab:fitting}, where our model achieves the best average scores across all metrics, including PSNR, SSIM~\cite{wang2004image}, and LPIPS~\cite{zhang2018unreasonable}, while being significantly more parameter-efficient than the second-best methods~\cite{shiinductive, rezaeian2025sl2a}. As shown in Fig.~\ref{fig:fitting}, our method achieves faster convergence and better reconstruction of fine structural details. We attribute this performance advantage to the model's strong joint frequency selection and spatial localization capabilities. Comprehensive per-scene qualitative and quantitative results are provided in the \textit{Supplementary Material}.

\noindent \textbf{3D Shape Representation.} Signed distance function (SDF) represents the distance from any 3D point to the closest surface using a continuous function, where the sign indicates whether the point lies inside (negative) or outside (positive) the object boundary~\cite{jones20063d}. We follow the baseline setup ~\cite{lindell2022bacon} and evaluate on five shapes from the Stanford 3D Scanning Repository~\cite{stanford3d}. Per-scene quantitative results in Table~\ref{tab:sdf} show that FLAIR outperforms competing approaches. As shown in Fig.~\ref{fig:SDF}, FLAIR also reconstructs fine-grained geometry with sharper and more faithful surface details than other methods.

\begin{table}[t]
\centering
\vspace{-0.2cm}
\caption{\textbf{Average results on the NeRF synthetic dataset}~\cite{mildenhall2021nerf}. 
Per-scene results are provided in the Supplementary.}
\vspace{-0.3cm}
\label{tab:nerf}
\scriptsize
\setlength{\tabcolsep}{3pt}
\renewcommand{\arraystretch}{1.1}
\resizebox{\linewidth}{!}{%
\begin{tabular}{lccccc}
\toprule
Metrics & IGA (ICML’25) & FINER (CVPR’24) & WIRE (CVPR’23) & SIREN (NeurIPS’20) & \textbf{FLAIR (Ours)} \\
\midrule
PSNR$\uparrow$   & 28.52 & \second{28.97} & 27.20 & 25.60 & \best{29.31} \\
SSIM$\uparrow$   & 0.9217 & \second{0.9261} & 0.9009 & 0.8857 & \best{0.9325} \\
LPIPS$\downarrow$& 0.063 & \second{0.044} & 0.065 & 0.105 & \best{0.041} \\
\bottomrule
\end{tabular}}
\end{table}

\subsection{Inverse Problems of 2D Images}
\label{sec:Inverse}
\noindent \textbf{Arbitrary-Scale Super-Resolution.}
INRs inherently act as continuous interpolants, enabling super-resolution beyond fixed scales. Accordingly, we evaluate our method under arbitrary-scale settings (\textit{e.g.}, $\times$4, $\times$6, $\times$8), with further results in the \textit{Supplementary}. As shown in Fig.~\ref{fig:SR}, FLAIR achieves the most fine-grained reconstruction at $\times$4, particularly around the butterfly’s head (red box).

\noindent \textbf{Image Denoising.}
Denoising is a particularly challenging problem that requires not only effective noise suppression but also the preservation of fine details. To address this challenge, precise frequency selection in regions where noise and structure interact is essential. FLAIR meets this requirement by effectively removing noise while preserving fine textures, as shown in the \textit{Supplementary}.

\begin{figure} \centering \includegraphics[width=\linewidth,keepaspectratio]{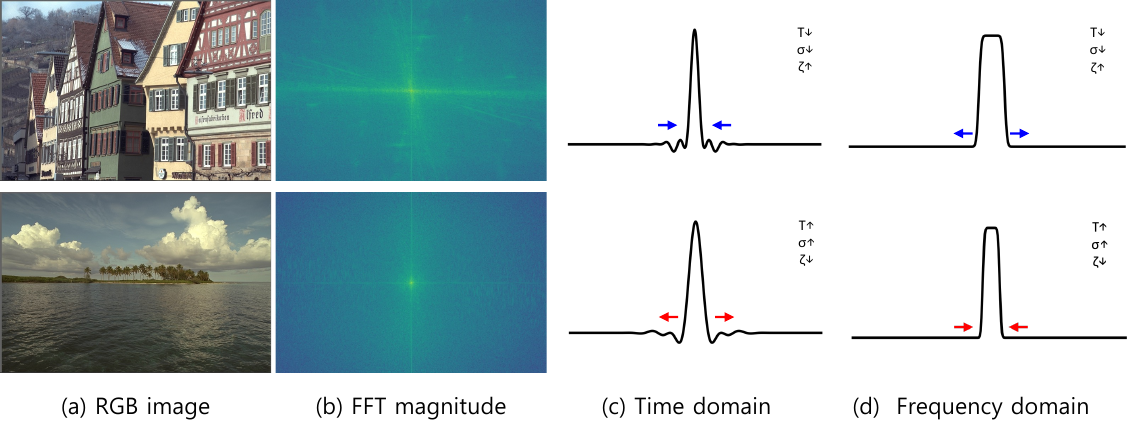}
\vspace{-0.7cm}
\caption{\textbf{Ablation study on scene adaptivity.} Comparison between a complex scene (top row) and a homogeneous scene (bottom row).}
\vspace{-0.2cm}
\label{fig:scene} 
\vspace{-0.1cm}
\end{figure}

\subsection{Neural Radiance Fields}
\label{sec:NeRF} 
\textbf{Neural Radiance Fields.} NeRFs~\cite{mildenhall2021nerf} synthesize novel views of complex 3D scenes by learning a continuous volumetric radiance field that maps 3D coordinates and viewing directions to color and density.
Following prior work~\cite{saragadam2023wire,liu2024finer}, we adopt a challenging setting using only 25 training images, instead of 
the standard 100. As shown in Fig.~\ref{fig:NeRF}, FLAIR remains robust under this limited-view regime and preserves structural details.
Table~\ref{tab:nerf} reports results on the Blender benchmark~\cite{mildenhall2021nerf}, where FLAIR achieves the best average performance across scenes. Additional per-scene results are provided in the \textit{Supplementary Material}.

\begin{table}[t]
\centering
\vspace{-0.2cm}
\caption{\textbf{Layer-wise parameters in BLA.} Each parameter ($T$, $\sigma$, $\zeta$) is a learned scalar, and the columns list the values learned in each layer $l$.}
\vspace{-0.3cm}
\label{table:layerwise}
\scriptsize
\renewcommand{\arraystretch}{0.85} 
\setlength\tabcolsep{5pt}
\resizebox{\linewidth}{!}{%
\begin{tabular}{l|c|cccc|c}
\toprule

\multirow{2}{*}{\parbox[c]{2.8cm}{\centering Scene}}& \multirow{2}{*}{Parameter} 
& \multicolumn{4}{c|}{Layer index $l$} & \multirow{2}{*}{\textbf{Average}} \\
\cline{3-6}
 &  & \rule{0pt}{1.1em}1 & 2 & 3 & 4 &  \\
\midrule

\multirow{3}{*}{\parbox[c]{2.8cm}{\centering Complex\\(Kodak 08)}}
& $T$      & 0.713 & 1.024 & 1.046 & 0.674 & \textbf{0.864} \\
& $\sigma$ & 1.955 & 2.011 & 2.041 & 1.468 & \textbf{1.868} \\
& $\zeta$  & 1.178 & 1.005 & 0.993 & 1.166 & \textbf{1.086} \\
\midrule

\multirow{3}{*}{\parbox[c]{2.8cm}{\centering Homogeneous\\(Kodak 16)}}
& $T$      & 1.033 & 1.026 & 0.962 & 1.008 & \textbf{1.005} \\
& $\sigma$ & 1.964 & 2.016 & 2.136 & 2.010 & \textbf{2.015} \\
& $\zeta$  & 1.114 & 0.999 & 0.989 & 1.021 & \textbf{1.031} \\
\bottomrule
\end{tabular}
}
\end{table}

\subsection{Ablation Study}
\label{sec:ablation}
\noindent \textbf{Scene-Adaptive Behavior of BLA.}
To validate whether the learnable parameters of BLA adapt to scene complexity, we conduct an image fitting experiment on two representative scenes, a complex scene (top row) and a homogeneous scene (bottom row), as shown in Fig.~\ref{fig:scene}. We fit BLA to each image and report the learned parameters \( (T, \sigma, \zeta) \) across layers in Table~\ref{table:layerwise}. The complex scene exhibits smaller average values of \(T\) and \(\sigma\), indicating stronger time-domain localization and broader frequency-domain support, as reflected in the top row (c) and (d) of Fig.~\ref{fig:scene}. In addition, its larger \(\zeta\) induces a stronger spectral shift observed in Fig.~\ref{fig:scene} (d), revealing increased emphasis on high-frequency components compared to the bottom row.

\noindent \textbf{Hyperparameter Analysis of WEGE.} As shown in Table~\ref{tab:wege}, we adopt the Daubechies-3 wavelet basis for its smoothness and compact support, and we vary the DWT level $J$. Among all settings, $J\!=\!1$ performs best, effectively preserving fine structures. We also examine the guided filter, which smooths edge discontinuities in the raw score and yields more stable representations. We set \(r=6\) and \(\epsilon=1\mathrm{e}{-5}\) to balance excessive sharpness and blur, reducing score discontinuities and preserving score accuracy. Additional score maps are provided in the \textit{Supplementary}.



\begin{table}[t]
\centering
\vspace{-0.2cm}
\caption{\textbf{Ablation of WEGE.} We evaluate the effect of the DWT level $J$ and the guided filter parameters (window radius $r$ and regularization $\epsilon$) within WEGE. Fitting experiments are conducted on five randomly selected DIV2K scenes (0, 5, 8, 9, 13). We report the average performance for brevity, while the relative ordering across individual scenes and various tasks remains consistent, indicating that the chosen configuration maintains generality.}
\vspace{-0.2cm}
\label{tab:wege}
\renewcommand{\arraystretch}{1.2}
\setlength\tabcolsep{4.5pt}
\resizebox{\linewidth}{!}{%
\begin{tabular}{c c c c c c}
\toprule
\multicolumn{3}{c}{ } & J = 1 & J = 2 & J = 3 \\
WEGE & \boldmath{$r$} & \boldmath{$\epsilon$} & PSNR$\uparrow$ / SSIM$\uparrow$ / LPIPS$\downarrow$ & PSNR$\uparrow$ / SSIM$\uparrow$ / LPIPS$\downarrow$ & PSNR$\uparrow$ / SSIM$\uparrow$ / LPIPS$\downarrow$  \\
\midrule
$\times$ & $\times$ & $\times$ &
\multicolumn{3}{c}{33.26 / 0.9479 / 0.050 \textit{(baseline without DWT)}} \\
\checkmark & $\times$ & $\times$
& 33.48 / 0.9494 / 0.031
& 30.88 / 0.9171 / 0.037
& 30.61 / 0.9152 / 0.046 \\
\checkmark & 2 & 1e-8
& 35.11 / 0.9659 / \second{0.020}
& 33.58 / 0.9540 / 0.033
& 33.53 / 0.9527 / 0.036 \\
\checkmark & 6 & 1e-5
& \best{36.39} / \best{0.9779} / \best{0.014}
& 34.92 / 0.9635 / 0.033
& 34.36 / 0.9591 / 0.035 \\
\checkmark & 10 & 1e-2
& \second{36.01} / \second{0.9717} / 0.023
& 34.14 / 0.9595 / 0.045
& 32.95 / 0.9431 / 0.040 \\
\bottomrule
\end{tabular}
}
\end{table}

\section{Conclusion}
We have proposed FLAIR, a Frequency- and Locality-Aware Implicit Neural Representation framework that unifies two complementary components. BLA is a novel activation function enabling precise frequency selection while suppressing excessive spatial oscillations, thereby facilitating effective signal modeling. WEGE is a lightweight module tailored to INR architectures, adding only 0.1K parameters. Our method consistently alleviates spectral bias and demonstrates superior fine-detail reconstruction across various tasks and datasets.

\section{Acknowledgement}

This work was supported by the Ministry of Education of the Republic of Korea and the National Research Foundation of Korea (NRF-2025S1A5C3A04022639). This research was supported by the Culture, Sports and Tourism R\&D Program through the Korea Creative Content Agency grant funded by the Ministry of Culture, Sports and Tourism in 2024 (Project Name: Developing Professionals for R\&D in Contents Production Based on Generative AI and Cloud, Project Number: RS-2024-00352578, Contribution Rate: 50\%). This research was supported by the MSIT (Ministry of Science and ICT), Korea, under the Graduate School of Virtual Convergence support program (IITP-2024-RS-2024-00418847) supervised by the IITP (Institute for Information \& Communications Technology Planning \& Evaluation).
{
    \small
    \bibliographystyle{ieeenat_fullname}
    \bibliography{main}
}
\clearpage
\setcounter{page}{1}
\maketitlesupplementary

\appendix
\vspace{-0.2cm}
In this supplementary material, we first present the proposed Band-Localized Activation (BLA) in Sec.~\ref{sec:sup_bla}, with ablations on its term components and initialization. Subsequently, in Sec.~\ref{sec:sup_wege}, we introduce the Wavelet-Energy-Guided Encoding (WEGE) as a plug-and-play module compatible with other activations. We further perform ablations on its guided-filter score-map hyperparameters and present additional visualizations, including variants with different guided-filter settings and the no-WEGE variant. Additionally, we provide mathematical full derivations in Sec.~\ref{sec:sup_math} and per-scene results in Sec.~\ref{sec:sup_results}, including convergence efficiency analysis. Finally, Sec.~\ref{sec:sup_E} provides a comprehensive discussion of the applications and limitations of FLAIR.

\begin{figure}[t]
\centering
\vspace{-0.1cm}
\includegraphics[width=\linewidth,keepaspectratio]{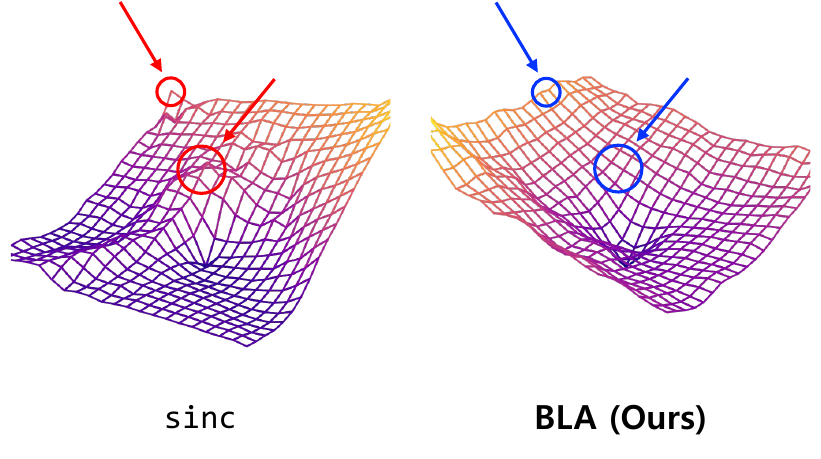}
\vspace{-0.7cm}
\caption{\textbf{Loss landscape visualizations on \texttt{sinc} and our proposed BLA.} Loss landscape visualizations show that Band-Localized Activation (BLA) produces a markedly flatter \textbf{MSE loss surface} than \texttt{sinc}. \textcolor{red}{\textbf{Red arrows}} indicate the sharp, irregular regions characteristic of \texttt{sinc}, whereas \textcolor{blue}{\textbf{blue arrows}} highlight the smoother, low-curvature regions formed by BLA, leading to more stable and reliable optimization.}
\label{fig:landscape}
\end{figure}

\section{Band-Localized Activation (BLA)}
\label{sec:sup_bla} 
\subsection{Loss Landscape Visualizations}

As shown in Fig.~\ref{fig:landscape}, loss landscape visualizations show that
Band-Localized Activation (BLA) has a flatter loss surface (MSE reconstruction)
than \texttt{sinc}. A flatter landscape is widely associated with improved
performance and better generalization~\cite{li2018visualizing, keskar2016large, foret2020sharpness}.
Thus, BLA improves not only accuracy but also robustness in large data regimes.

To obtain the loss landscapes, we follow the perturbation-based analysis of
Li~\textit{et al.}~\cite{li2018visualizing}. Specifically, given the optimized
parameters $\theta$, we sample two random filter-normalized directions
$d_{1}$ and $d_{2}$ in the weight space. We then evaluate the perturbed models 
$\theta'=\theta+\alpha d_{1}+\beta d_{2}$ 
over a 2D grid of $(\alpha,\beta)$ values, where both variables are uniformly sampled within a fixed range $[-1,1]$ to form an $n\times n$ lattice of 
perturbations. For each lattice point, we compute the MSE reconstruction loss, 
yielding a faithful 2D slice of the loss landscape that enables a clear 
comparison of sharpness and stability.

\subsection{Theoretical Foundations of Signal Reconstruction}

As introduced in Sec.~\ref{sec:riesz}, Riesz signal reconstruction provides a general framework for stable signal representation.
A Riesz basis must satisfy two conditions: the Riesz bounds and the partition-of-unity (PUC).
Our BLA is constructed to meet both conditions, and therefore Riesz basis theory becomes the theoretical foundation for BLA. Furthermore, this theoretical foundation provides the basis for the BLA initialization scheme described in Sec.~\ref{sec:bla_init}.

\subsection{Full Derivation of BLA}

Within the Riesz basis framework (Sec.~\ref{sec:riesz}), both the sinc function and our BLA satisfy the required Riesz bounds and partition-of-unity conditions. In deep learning optimization, however, the residual term in Eq.~\eqref{eq:oscillation} is unavoidable, and this oscillatory component induces instability during training. This behavior is reflected in the perturbation experiments of Fig.~\ref{fig:landscape}, where sinc exhibits a distinctly non-flat loss landscape. The theoretical differences are formalized in the derivations of Sec.~\ref{sec:sinc_grad} and Sec.~\ref{sec:bla_grad}. Practically, the effect is evident in the MSE convergence of Fig.~\ref{fig:fitting}(a) and in Table~\ref{tab:bla_ablation}, where BLA consistently improves over the pure band-limiting (BL) baseline across all metrics.
\begin{table}[t]
\centering
\caption{
\textbf{Ablation of the proposed BLA activation and the WEGE module.}
Columns progressively activate the four components of BLA:
band-limiting (BL), transition-smoothing (TS), localization (LOC),
and frequency-shifting (FS), forming the full BLA
\mbox{(BL × TS × LOC × FS)}. The final column additionally applies
WEGE on top of BLA, corresponding to the full FLAIR model. All values are evaluated on DIV2K~\cite{agustsson2017ntire} Image~00.}
\vspace{-0.2cm}
\label{tab:bla_ablation}
\renewcommand{\arraystretch}{1.15}
\setlength{\tabcolsep}{6pt}

\resizebox{\linewidth}{!}{%
\begin{tabular}{l|c|c|c|c|c}
\hline
& \textbf{BL (base)} & \textbf{+ TS} & \textbf{+ LOC} &
\textbf{BLA (full)} & \textbf{BLA + WEGE} \\
\hline
PSNR $\uparrow$      & 27.31 & 28.18 & 30.22 & 32.96 & \textbf{34.75} \\
SSIM $\uparrow$      & 0.8243 & 0.8581 & 0.9069 & 0.9380 & \textbf{0.9663} \\
LPIPS $\downarrow$   & 0.101 & 0.087 & 0.065 & 0.034 & \textbf{0.013} \\
\hline
\end{tabular}
}
\end{table}
\subsection{Ablation of BLA Components and WEGE Integration}

In Table~\ref{tab:bla_ablation}, each component yields consistent improvements.
TS and LOC yield notable spatial localization gains. Moreover, the interaction
between FS and WEGE provides a strong and complementary synergy, enabling more
accurate modeling of the target signal.

\clearpage

\begin{figure*}[t]
\centering
\includegraphics[width=\linewidth,keepaspectratio]{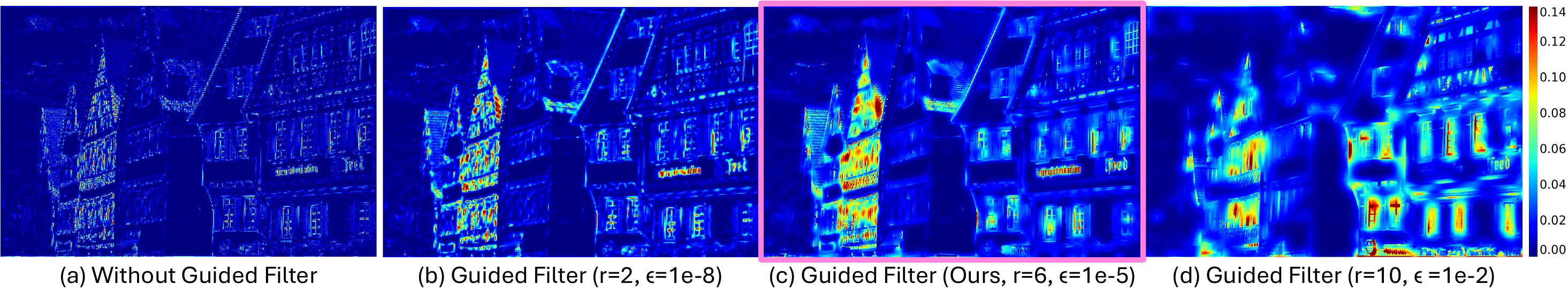}
\vspace{-0.7cm}
\caption{\textbf{Guided filtering effects on WEGE score-map visualizations.} \(r\) is the window radius and \(\epsilon\) is the regularization parameter, and larger \(r\) together with larger \(\epsilon\) yields a smoother filter in (d).}
\label{fig:score}
\end{figure*}

\vspace{0.5cm}
\begin{figure*}[t]
\centering
\includegraphics[width=\linewidth,keepaspectratio]{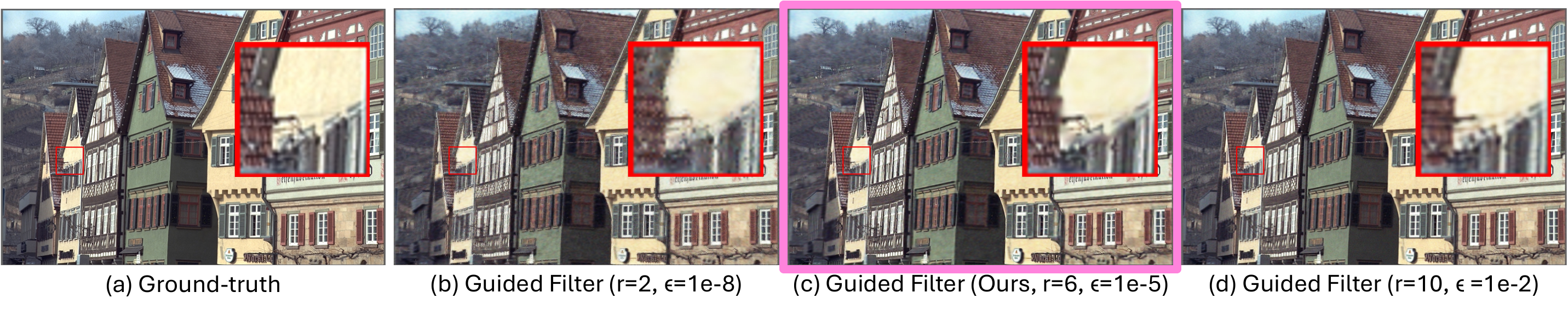}
\vspace{-0.6cm}
\caption{\textbf{Qualitative effects of guided filtering and WEGE in the RGB domain.}  (c) follows our guided-filter setting (\(r=6\), \(\epsilon=1\mathrm{e}{-5}\)), preserving fine details without noise and is closest to (a) ground truth.}
\label{fig:rgb}
\end{figure*}

\section{Wavelet-Energy-Guided Encoding (WEGE)}
\label{sec:sup_wege}  
\vspace{0.5cm}
\subsection{Effect of Guided Filtering on WEGE Scores}

In Fig.~\ref{fig:score}, we present score maps produced with different radius \(r\) and regularization \(\epsilon\) settings to clarify our chosen hyperparameters, and we show their resulting effects on the final RGB outputs in Fig.~\ref{fig:rgb}.

As shown in Fig.~\ref{fig:score}, (a) presents the raw WEGE score without guided filtering, while (b)–(d) apply guided filtering with different radius \(r\) and regularization \(\epsilon\). Smaller \(r\) and \(\epsilon\) preserve sharper local structures, whereas larger \(r\) and \(\epsilon\) progressively smooth the score responses. The score discontinuities produced by sharp filtering lead to salt-and-pepper–like artifacts, as shown in Fig.~\ref{fig:rgb}(b), while overly smoothed scores result in blurry roof boundaries, as shown in Fig.~\ref{fig:rgb}(d). In contrast, the chosen hyperparameters in (c) yield a balanced score representation that preserves fine details while suppressing artifacts. This choice is further validated across diverse scenes and tasks by the quantitative results in Table~\ref{tab:wege} of the main paper.

\subsection{Plug-and-Play Behavior of WEGE}

\begin{figure} 
\vspace{-0.3cm}
\centering \includegraphics[width=\linewidth,keepaspectratio]{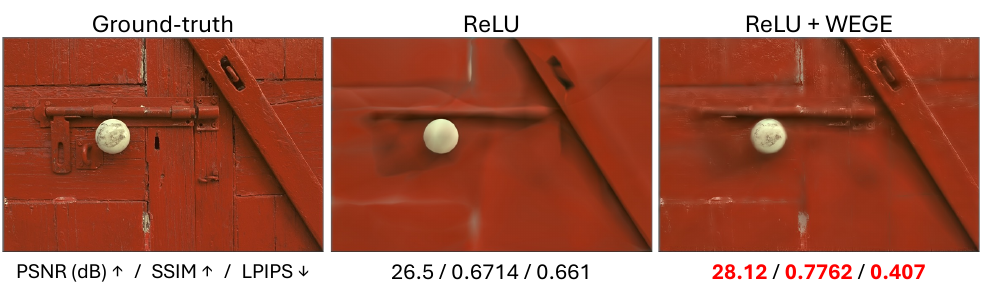}
\vspace{-0.7cm}
\caption{\textbf{Plug-and-play WEGE on ReLU.} WEGE substantially improves both the visual and quantitative quality of ReLU-based INRs.}
\label{fig:WEGE_fitting} 
\vspace{-0.2cm}
\end{figure}

\begin{figure} \includegraphics[width=\linewidth,keepaspectratio]{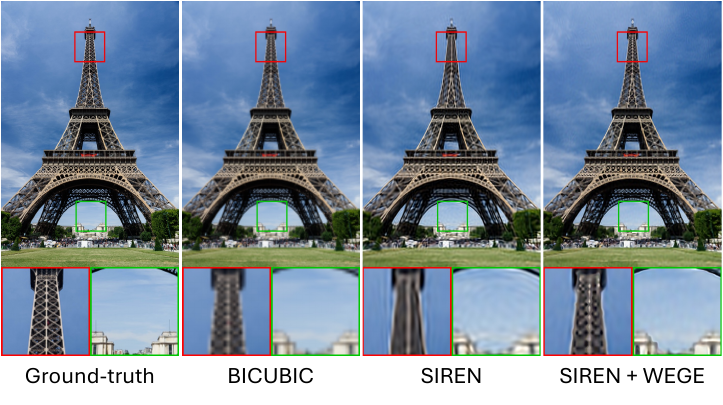}
\vspace{-0.7cm}\centering
\caption{\textbf{Plug-and-play WEGE on SIREN for x4 super-resolution.} WEGE enhances SIREN’s fine details (red box) while suppressing noise (green box) through explicit frequency guidance.}
\label{fig:WEGE_SR} 
\vspace{-0.5cm}
\end{figure}

We evaluate WEGE in a plug-and-play setting by integrating it into baseline activations like ReLU and SIREN. 
Fig.~\ref{fig:WEGE_fitting} shows that WEGE enhances fine-detail reconstruction for the image fitting task and provides consistent improvements in PSNR, SSIM~\cite{wang2004image}, and LPIPS~\cite{zhang2018unreasonable}.
These results indicate that WEGE refines the embedding space from \(f(x,y)\) to \(f(x,y,\tilde{w}_b)\) through its frequency guidance, enabling more expressive and precise structural representations. Beyond fitting, WEGE extends to diverse tasks. 
Fig.~\ref{fig:WEGE_SR} demonstrates its effectiveness on the super-resolution task, where frequency-aware refinement mitigates frequency leakage and improves high-frequency structure reconstruction.

\clearpage
\section{Theoretical Foundations for BLA}
\label{sec:sup_math}

\subsection{Classical Sampling Theory}
\label{sec:classic}
The primary objective of classical sampling theory is the reconstruction of a continuous signal from its discrete samples. This objective conceptually aligns with Implicit Neural Representations (INRs), which map low-dimensional coordinates to signal values while learning a continuous representation from discrete data. 

Classical sampling theory specifies when a bandlimited signal can be perfectly recovered from its discrete samples. A signal $f$ is $T$-bandlimited if its Fourier transform $\hat{f}(s)$ satisfies $\hat{f}(s)=0$ for all $|s|>T$. This definition makes the notion of perfect reconstruction well posed. In particular, the Nyquist-Shannon sampling theorem establishes that a $T$-bandlimited signal $f \in L^2(\mathbb{R})$ can be perfectly reconstructed as:
\begin{equation}
    \small
    f(x) = \sum_{n=-\infty}^{\infty} f\left(\frac{n}{2T}\right) \texttt{sinc}\left(2T x - n\right),
\end{equation}
where $T$ denotes the maximum frequency of the signal $f(x)$. The quantity $2T$ is the Nyquist sampling rate, defined as twice this maximum frequency. Its reciprocal, $1/(2T)$, specifies the sampling interval, and the discrete samples $f(n/(2T))$ serve as the coefficients used in the reconstruction formula. The term \texttt{sinc}$(2Tx - n)$ denotes the sinc interpolation kernel scaled by the Nyquist rate $2T$ and shifted according to the sample index $n$.

In practice, however, classical sampling theory faces inherent limitations due to the characteristics of real-world signals. Although it guarantees theoretically perfect reconstruction under the bandlimited assumption, achieving this requires an infinite number of samples, which is infeasible in practical settings. These limitations necessitate a more general signal reconstruction framework, for which Riesz basis theory provides the appropriate mathematical framework.

\subsection{Riesz Basis Theory}
\label{sec:riesz}
Riesz basis theory provides a generalized mathematical framework that overcomes the practical limitations of classical sampling theory. In this context, the Nyquist–Shannon sampling theorem, which represents a specific instance of classical sampling theory, 
is limited to the \texttt{sinc} function and strictly bandlimited signals. In contrast, Riesz basis theory generalizes this framework by enabling reconstruction with a much broader class of basis functions. The core concept of the Riesz basis is to specify the conditions under which the set of shifted functions $\{F(x-k)\}_{k\in\mathbb{Z}}$ forms a stable and complete representation system within a specified signal space.

Specifically, the Riesz basis framework defines the $T$-scaled signal space for $T>0$ as follows:
\begin{equation}
\scalebox{0.85}{%
$\displaystyle
V_T(F) = \left\{ s_T(x) = \sum_{k \in \mathbb{Z}} a_T(k)\,
F\!\left(\frac{x}{T} - k\right) : a_T \in l^2(\mathbb{R}) \right\},$
}
\end{equation}
where $T$ determines the spatial scale of the signal space. As $T$ decreases, the basis function $F$ becomes spatially compressed, enabling the representation of higher-frequency components.  
Conversely, larger values of $T$ expand $F$, capturing lower-frequency structures.  
Here, $s_T(x)$ denotes an arbitrary signal belonging to the space $V_T(F)$. This signal is expressed as a linear combination of the basis functions $F\left(\frac{x}{T} - k\right)$. The term $a_T(k)$ represents the $k$-th coefficient multiplied by each basis function, where the sequence $a_T$ is an element of $\ell^2(\mathbb{R})$, the Hilbert space of square-summable real sequences.

In particular, for the $T$-scaled signal space $V_T(F)$ to operate as a signal reconstruction system, the set of shifted functions $\{F_k = F(x-k)\}_{k\in\mathbb{Z}}$ must form a Riesz basis. A Riesz basis must satisfy the following two conditions:
\begin{align}
\text{1.}\quad 
& \scalebox{0.9}{$
A\|a\|_{l^2}^2 
  \le 
  \left\| \sum_{k\in\mathbb{Z}} a(k)F_k \right\|^2 
  \le 
  B\|a\|_{l^2}^2,
\ \forall a(k)\in\!\ell^2(\mathbb{R}).
$}
\label{eq:riesz bounds}
\\[0.9em]
\text{2.}\quad 
& \scalebox{0.9}{$
\sum_{k\in\mathbb{Z}} F(x+k) = 1,
\ \forall x\in\mathbb{R}.
$}
\label{eq:partition of unity}
\end{align}
The first condition ensures a stable representation for the linear combination of the basis functions, where $A$ and $B$ are positive constants satisfying $0 < A \leq B < \infty$. The lower bound of this inequality guarantees the linear independence of the basis functions $\{F_k\}_{k \in \mathbb{Z}}$. The upper bound ensures that the $L^2$-norm of any signal $s_T \in V_T(F)$ is finite, thereby establishing $V_T(F)$ as a subspace of $L^2(\mathbb{R})$. 

The second condition, known as the Partition of Unity Condition (PUC), ensures that the Riesz basis can approximate general \(L^2(\mathbb{R})\) 
signals beyond the space \(V_T(F)\). When this condition is satisfied, selecting an appropriate scale $T$ enables the approximation of arbitrary $L^2(\mathbb{R})$ signals with arbitrary precision.

Riesz basis theory guarantees that a set of shifted basis functions $\{F(x-k)\}_{k\in\mathbb{Z}}$ that forms a Riesz basis can approximate any signal in $L^2(\mathbb{R})$ with arbitrary precision. Specifically, for any signal $s \in L^2(\mathbb{R})$ and any arbitrarily small positive value $\epsilon > 0$, there exists an appropriate scale $T > 0$ and a function $f_T \in V_T(F)$ such that the approximation error is bounded by $|s - f_T|_{L^2} < \epsilon$. This guarantee is attributed to the PUC, which performs a critical role in bridging the gap between the $T$-scaled space $V_T(F)$ and the general $L^2(\mathbb{R})$ space.

\subsection{Gradient Derivation of Sinc-activated INR}
\label{sec:sinc_grad}
The $\texttt{sinc}$ function is a canonical example of a function that forms a Riesz basis, satisfying the two required conditions. This theoretical foundation suggests that the $\texttt{sinc}$ function is sufficient for optimal signal reconstruction. However, this theoretical guarantee establishes only the existence of a function $f_T \in V_T(F)$ that can approximate any signal in $L^2(\mathbb{R})$ with arbitrary precision, and it does not provide a specific methodology for practically obtaining the $f_T$. In particular, when the $\texttt{sinc}$ function is employed as the activation function within the INR framework for signal reconstruction, its global oscillatory property can induce training instability. This instability can be mathematically analyzed by deriving the gradient of the $\texttt{sinc}$ function.

The output of the INR network $f(x; W)$ for a single input coordinate $x \in \mathbb{R}^{d_{\text{in}}}$ is expressed as follows, omitting biases and the output layer weights for simplification:
\begin{equation}
    \small
    f(x; W) = \sum_{i=1}^{d_{\text{out}}} \frac{1}{T} F\left(\frac{W_i x}{T}\right),
\end{equation}
where $W \in \mathbb{R}^{d_{\text{out}} \times d_{\text{in}}}$ is the weight matrix of the hidden layer, $F(\cdot)$ is the activation function, and $T > 0$ represents the frequency scaling parameter. The loss function utilized for signal reconstruction is defined as the $L^2$ loss between the target signal $g$ and the network output $f$:
\begin{equation}
    \small
    \mathcal{L}(W) = \|g - f\|_{2}^2 = \int |g(x) - f(x; W)|^2 dx.
\end{equation}
For notational simplicity, the continuous representation is employed in the subsequent derivation instead of the discrete representation. The gradient of the loss function with respect to the weight $W_{ij}$ is derived by applying the chain rule, as follows:
\begin{equation}\label{eq:gradient derivation}
    \small
    \frac{\partial \mathcal{L}}{\partial W_{ij}} = -2 \int [g(x) - f(x; W)] \cdot \frac{\partial f(x; W)}{\partial W_{ij}} dx.
\end{equation}
Thus, the gradient of the INR using the $\texttt{sinc}$ activation function is derived as follows:
\begin{equation}
    \small
    \frac{\partial \mathcal{L}}{\partial W_{ij}} = -\frac{2}{\pi} \int \frac{g(x) - f(x)}{(W_i x)^2} \left[\pi u \cos(\pi u) - \sin(\pi u)\right] x_j \, dx,
\end{equation}
where $u = W_i x / T$, and $x_j \in \mathbb{R}$ is the value corresponding to the $j$-th dimension. The derived gradient expression indicates two major properties that contribute to the training instability of INRs using the $\texttt{sinc}$ activation function. The derived gradient expression explains two major causes of training instability in $\texttt{sinc}$-activated INRs.

First, the term $[\pi u \cos(\pi u) - \sin(\pi u)]$ exhibits oscillatory behavior with respect to $u = W_i x / T$. This behavior causes the gradient's sign to frequently change across the input space, leading to an unstable convergence path during optimization. Second, the $\texttt{sinc}$ function is defined as $\texttt{sinc}(x) = \sin(x) / x$ and exhibits non-zero values theoretically over an infinite range. Consequently, the gradient for each weight $W_{ij}$ is influenced by the error $[g(x) - f(x)]$ across the entire input domain.

This analysis suggests the motivation for establishing an activation function that can ensure practical training stability while preserving the theoretical guarantees of the Riesz basis. Specifically, functions with compact support have non-zero values only within a limited spatial region, a property which allows them to effectively mitigate the issues of global oscillation and residual error propagation resulting from the global support of the $\texttt{sinc}$ function.

\subsection{Gradient Derivation of BLA-activated INR}
\label{sec:bla_grad}
To ensure training stability while maintaining the theoretical guarantees of the Riesz basis, we propose a novel activation function, the Band-Localized Activation (BLA), which incorporates a localization term. The gradient of the BLA is derived as follows, based on Eq.~\eqref{eq:gradient derivation}:
\begin{equation}
    \small
    \frac{\partial \mathcal{L}}{\partial W_{ij}} = -\frac{2}{\pi} \int \frac{[g(x) - f(x)] \cdot x_j}{(W_i x)^2} \cdot \exp\left(-\frac{u^2}{\sigma^2}\right) \cdot H(u) \, dx,
\end{equation}

\begin{equation}
    H(u) = \pi u \cos(\pi u) - \sin(\pi u)\left(1 + \frac{2u^2}{\sigma^2}\right).
\end{equation}
This gradient expression clearly demonstrates two key mechanisms by which the BLA function enhances training stability compared to the $\texttt{sinc}$ function.

First, the exponential decay term $\exp\left(-\frac{u^2}{\sigma^2}\right)$ exponentially reduces the contribution from regions distant from the origin in the input space. This ensures that the gradient is dominated by the error in local regions, thereby effectively mitigating global residual error propagation. Second, the exponential decay term spatially restricts the overall magnitude of the gradient. This exponentially attenuates the effective contribution of the oscillatory term $H(x)$ on the gradient in regions distant from the origin. The oscillatory term itself remains, but its influence is restricted to the local domain, thus mitigating instability during the training process.

This gradient analysis shows that the BLA function ensures training stability while maintaining the theoretical signal reconstruction capability of the $\texttt{sinc}$ function. To formally establish the theoretical foundation of the BLA within the Riesz basis framework, the proof of whether BLA satisfies the Riesz basis conditions is required. This proof will enable establishing the strategy to initialize the parameters of the BLA.

\subsection{BLA Initialization Scheme under the Riesz Basis Framework}
\label{sec:bla_init}
The BLA function must satisfy the Riesz basis condition to guarantee the theoretical foundation for signal reconstruction. Specially, this requires simultaneous adherence to two essential conditions: the stability (Eq.~\eqref{eq:riesz bounds}) and the partition of unity condition (Eq.~\eqref{eq:partition of unity}).

To verify the stability condition (Eq.~\eqref{eq:riesz bounds}), we reformulate it as follows using the Poisson summation formula~\cite{stein2011fourier}:
\begin{equation}\label{eq:fourier riesz bounds}
    A \leq \sum_{k \in \mathbb{Z}} |\hat{F}(\xi + 2k\pi)|^2 \leq B,
\end{equation}
where the lower bound $A > 0$ guarantees the linear independence of the basis functions. The upper bound $B < \infty$ ensures numerical stability by preventing excessive overlap in the frequency domain. Here, $\hat{F}$ denotes the Fourier Transform of the BLA function and $\xi$ is the frequency variable. Subsequently, the Fourier transform $\hat{F}(\xi + 2 k \pi)$ of the BLA function is derived as follows:
\begin{equation}\label{eq:BLA poisson summation}
    \hat{F}(\xi + 2 k \pi) = \frac{\sigma}{\sqrt{2 \pi}} \int_{-\pi / T}^{\pi / T} e^{-\sigma^2(\xi + 2 k \pi - \zeta)^2/2} d\zeta.
\end{equation}
Subsequently, we use the following inequality to derive the lower bound $A$ in Eq.~\eqref{eq:BLA poisson summation}~\cite{stein2011fourier}:
\begin{equation}\label{eq:riesz 1 poisson summation}
    \sum_{k \in \mathbb{Z}} \left| \hat{F}(\xi + 2k\pi) \right|^2 \geq \left| \hat{F}(\xi) \right|^2.
\end{equation}
In this equation, $|\hat{F}(\xi)|$ serves as the lower bound of $|\hat{F}(\xi + 2k\pi)|$, which is $A$ in Eq.~\eqref{eq:fourier riesz bounds}. Furthermore, since the integrand in Eq.~\eqref{eq:BLA poisson summation} decays exponentially for $k \neq 0$, this ensures that the upper bound $B$ in Eq.~\eqref{eq:fourier riesz bounds} is finite. Consequently, the existence of $A$ and $B$ satisfying $0 < A \leq B < \infty$ verifies that the BLA function satisfies the stability condition (Eq.~\eqref{eq:fourier riesz bounds}).

We verify the Partition of Unity Condition (PUC). By the Poisson summation formula, Eq.~\eqref{eq:partition of unity} is reformulated as follows:
\begin{equation}\
    \sum_{n \in \mathbb{Z}} \hat{F}(2\pi n) e^{2\pi i n x} = 1.
\end{equation}
This condition requires that $\hat{F}(0) = 1$ for $n = 0$ and $\hat{F}(2\pi n) = 0$ for $n \neq 0$.

We now verify that the BLA function satisfies the PUC by expanding $\hat{F}(2\pi n)$. The Fourier transform $\hat{F}(2\pi n)$ of the BLA function can be expanded as follows:
\begin{equation}\label{eq:fourier puc 2}
    \hat{F}(2\pi n) = \frac{T\sigma}{\sqrt{2\pi}} \int_{-\pi/T}^{\pi/T} e^{-\sigma^2(2\pi n - \zeta)^2/2} d\zeta.
\end{equation}
Note that $\hat{F}(2\pi n)$ is a function of $T$ and $\sigma$. We now determine the range of $T$ and $\sigma$ satisfying $\hat{F}(0) = 1$ and $\hat{F}(2\pi n) = 0$ to prove that the BLA function satisfies the PUC. To this end, we compute $\hat{F}(0)$ from Eq.~\eqref{eq:fourier puc 2}:
\begin{equation}\label{eq:fourier puc 1}
\small
\hat{F}(0)
= \frac{T}{\sqrt{\pi}}
\int_{-\sigma\pi/(T\sqrt{2})}^{\sigma\pi/(T\sqrt{2})} e^{-u^2}\, du
= T \cdot \texttt{erf}\!\left(\frac{\sigma\pi}{T\sqrt{2}}\right)
= 1,
\end{equation}
where $u = \sigma \zeta / \sqrt{2}$, and $\texttt{erf}(\cdot)$ denotes the error function, which satisfies $|\texttt{erf}(\cdot)| < 1$.

We now derive the constraint requiring $\hat{F}(2\pi n) = 0$ for $n \neq 0$. Since the integral in Eq.~\eqref{eq:fourier puc 2} cannot be computed analytically for all $n \in \mathbb{Z}$, we derive its upper bound as follows:
\begin{equation}\label{eq:rest components}
    \hat{F}(2 \pi) = \sigma\sqrt{2\pi} \cdot e^{-\frac{1}{2} \sigma^2 (2 \pi - \frac{\pi}{T})^2} = 0.
\end{equation}
The convergence of Eq.~\eqref{eq:rest components} to zero directly ensures that $\hat{F}(2\pi n) \to 0$ for all $n \in \mathbb{Z}$ by definition of the upper bound.

We have derived two constraints for the BLA function to satisfy the PUC: Eq.~\eqref{eq:fourier puc 2} and Eq.~\eqref{eq:fourier puc 1}. Since each constraint depends on $T$ and $\sigma$, we define the following cost function for these constraints:
\begin{equation}\label{eq:riesz basis error}
    E(T, \sigma) = |1 - \hat{F}(0)| + |\hat{F}(2 \pi)|.
\end{equation}
The theoretical optimal values of $T$ and $\sigma$ for the BLA to form a Riesz basis through Eq.~\eqref{eq:riesz basis error}. From Eq.~\eqref{eq:fourier puc 1} with $|\texttt{erf}(\cdot)| < 1$, we obtain $T \geq 1$. To maximize the frequency bandwidth, we set $T=1$ since smaller $T$ values yield wider bandwidths. We then choose $\sigma = 2$, the smallest value at which the rate of decrease in Eq.~\eqref{eq:riesz basis error} becomes negligible, thereby maximizing both spatial localization and frequency bandwidth.

\clearpage
\section{Additional Results Across Diverse Tasks}
\label{sec:sup_results}

\subsection{Signal Representation}
\begin{figure*} \centering 
\includegraphics[width=\linewidth,keepaspectratio]
{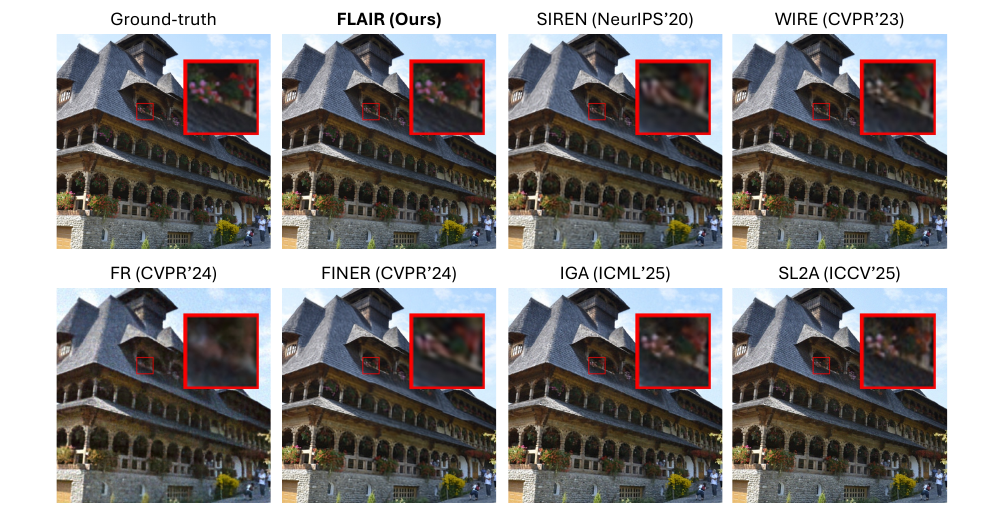} 
\caption{\textbf{Image fitting results on DIV2K image 13.}}
\label{fig:sup_div2k} 
\end{figure*}

\begin{figure*} \centering 
\includegraphics[width=\linewidth,keepaspectratio]
{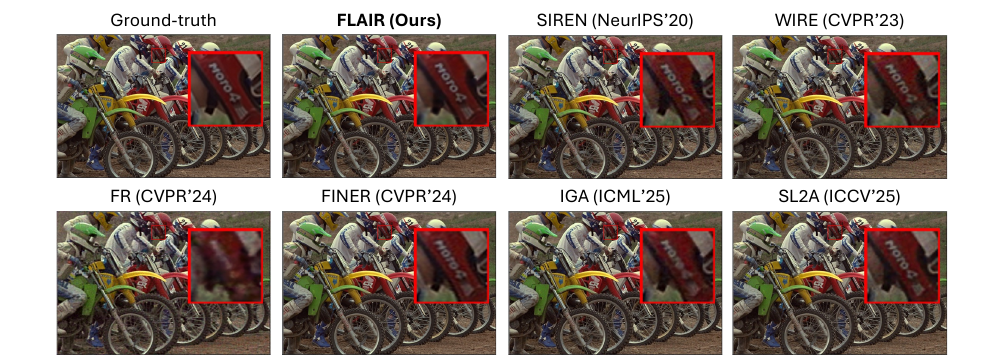} 
\caption{\textbf{Image fitting results on Kodak image 05.}}
\label{fig:sup_kodak} 
\end{figure*}

\textbf{2D Image Representation.} As shown in Figs.~\ref{fig:sup_div2k} and
\ref{fig:sup_kodak}, FLAIR produces superior per-scene reconstruction quality,
preserving fine details while modeling RGB values with higher fidelity and
minimal noise. Although Table~\ref{tab:fitting} in the main paper summarizes the
overall performance, we additionally provide full per-scene results for DIV2K and
Kodak in Tables~\ref{tab:div2k} and \ref{tab:kodak} for completeness.

\begin{figure*} \centering 
\includegraphics[width=\linewidth,keepaspectratio]
{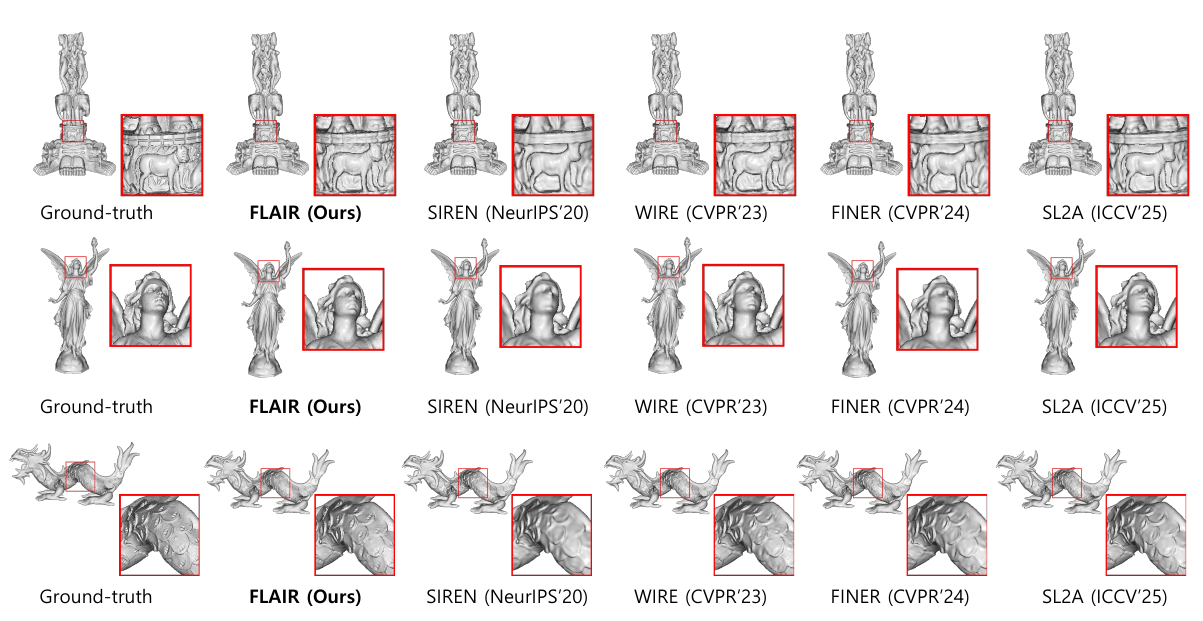} 
\caption{\textbf{Per-scene visual comparisons on signed distance field reconstruction.} 
FLAIR consistently outperforms other methods across all scenes, recovering finer geometry and sharper surface details.}
\label{fig:sup_SDf} 
\end{figure*}

\noindent{\textbf{3D Shape Representation.}} As shown in Fig.~\ref{fig:sup_SDf}, FLAIR
accurately reconstructs diverse 3D shapes, demonstrating robustness and
generality beyond data-specific biases. Together with our 2D results, FLAIR consistently outperforms prior methods across both 2D and 3D representation tasks, highlighting its strong scalability across modalities.
\begin{table}[t]
\centering
\caption{\textbf{Arbitrary-scale super-resolution comparison.}
Quantitative results on the Kodak dataset at $\times 6$ and $\times 8$ upscaling. \best{Red} and \second{blue} indicate best and second-best values.}
\label{tab:arbitrary_sr}

\renewcommand{\arraystretch}{1.15}
\setlength{\tabcolsep}{3.5pt}

\resizebox{\linewidth}{!}{%
\begin{tabular}{l|ccc|ccc}
\hline
\multirow{2}{*}{Methods} &
\multicolumn{3}{c|}{$\times 6$} &
\multicolumn{3}{c}{$\times 8$} \\
 & PSNR$\uparrow$ & SSIM$\uparrow$ & LPIPS$\downarrow$
 & PSNR$\uparrow$ & SSIM$\uparrow$ & LPIPS$\downarrow$ \\
\hline
FINER (CVPR’24)         & 23.94 & 0.6495 & 0.425 & 22.57 & 0.5761 & 0.549 \\
WIRE (CVPR’23)          & 22.29 & 0.4770 & \second{0.273} & 19.12 & 0.3191 & \second{0.392} \\
GAUSS (ECCV’22)         & \second{24.24} & \second{0.6601} & 0.382 & \second{22.84} & \second{0.6051} & 0.492 \\
MFN (ICLR’21)           & 20.64 & 0.3066 & 0.594 & 18.93 & 0.5761 & 0.667 \\
SIREN (NeurIPS’20)      & 24.18 & 0.6568 & 0.384 & 22.84 & 0.5801 & 0.443 \\
ReLU+P.E. (NeurIPS’20)  & 22.51 & 0.6073 & 0.398 & 21.85 & 0.5870 & 0.415 \\
\hline
\textbf{FLAIR (Ours)}   & \best{24.32} & \best{0.6786} & \best{0.246} & \best{23.92} & \best{0.6365} & \best{0.339} \\
\hline
\end{tabular}}
\end{table}

\subsection{Image Restoration Tasks}
\begin{figure*}
\centering
\includegraphics[width=\linewidth,keepaspectratio]{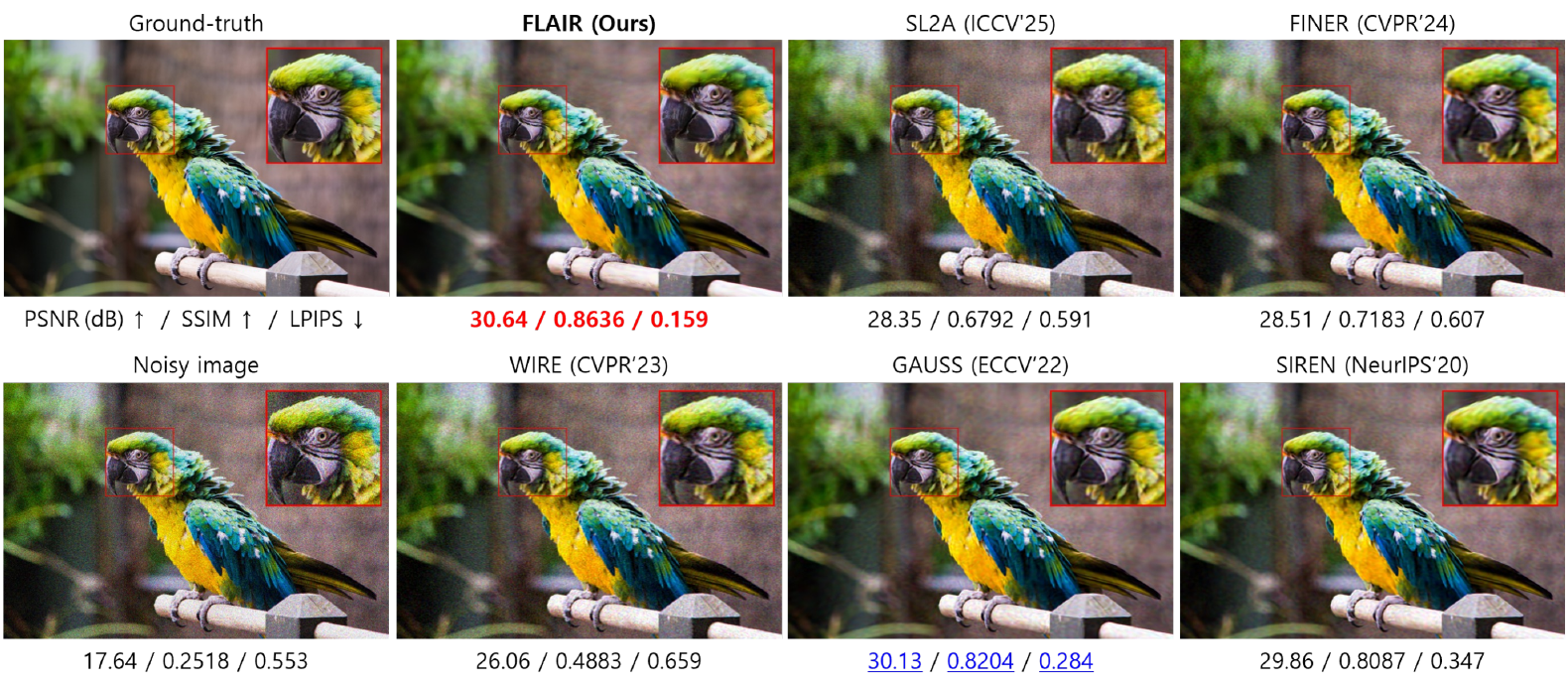}
\caption{\textbf{Visual comparisons between FLAIR and baselines on the denoising task.} FLAIR outperforms the baselines by effectively suppressing noise while preserving fine details.}
\label{fig:sup_denoising}
\end{figure*}

\textbf{Image Denoising.}
Image denoising is challenging because effective noise removal must be achieved
without degrading fine structures. As shown in Fig.~\ref{fig:sup_denoising},
FLAIR achieves both strong noise suppression and faithful detail preservation. FLAIR achieves both by leveraging the joint effect of band-limiting and spatial localization, rather than merely focusing on expressiveness as other baselines do.

\noindent{\textbf{Arbitrary-Scale Super-Resolution.}} 
Conventional SR networks rely on scale-specific upsampling modules, which constrain them to predefined output resolutions. In contrast, INRs represent signals as continuous coordinate-based mappings, inherently operating as resolution-independent interpolators. This enables super-resolution beyond fixed scales and supports seamless generation at arbitrary resolutions. Leveraging this capability, we demonstrate that FLAIR maintains strong performance even at extreme scales. As shown in Table~\ref{tab:arbitrary_sr}, FLAIR consistently outperforms existing methods across challenging upscaling factors.

\subsection{Neural Radiance Fields }
\begin{figure*}
\centering
\includegraphics[width=\linewidth,keepaspectratio]
{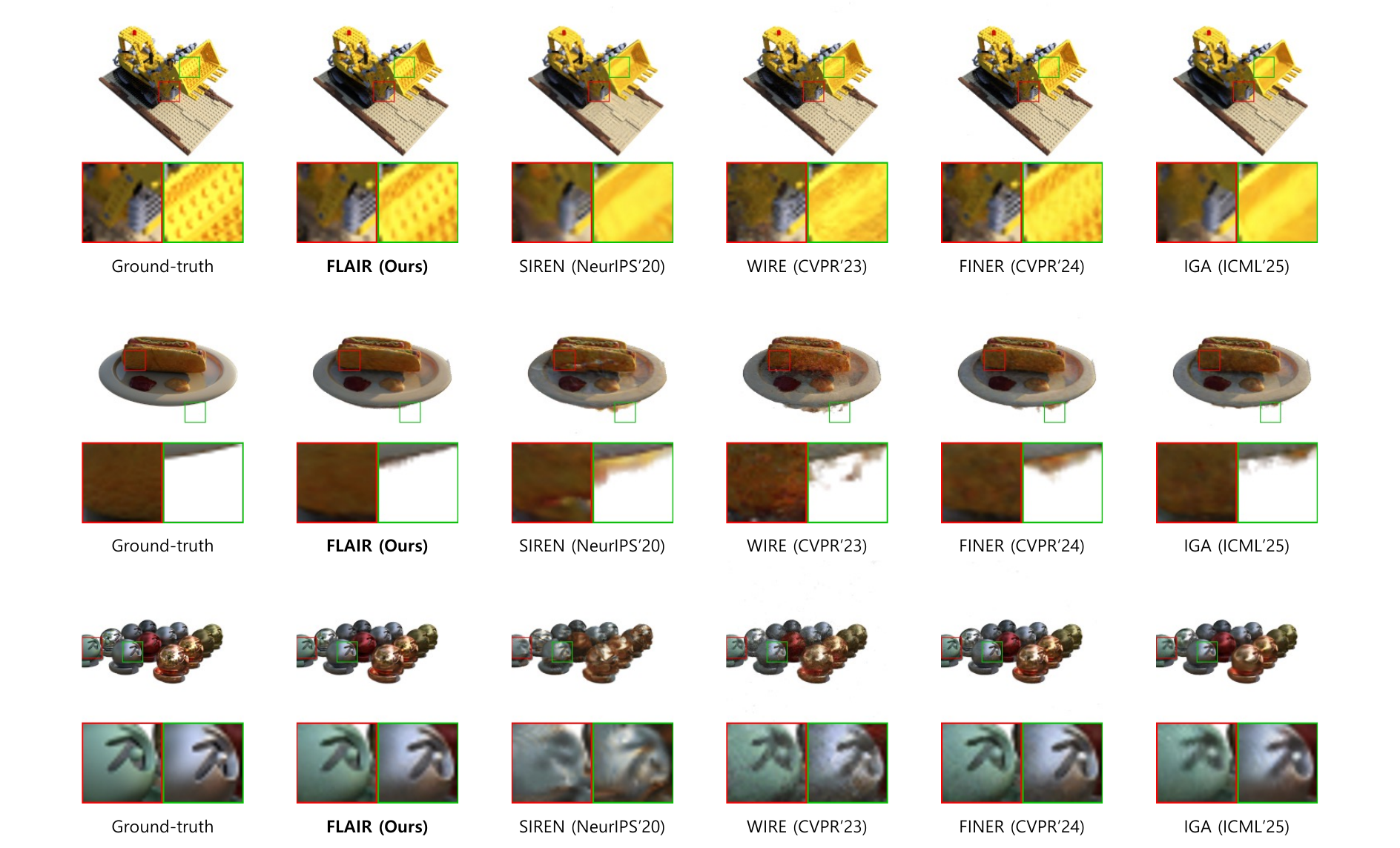}
\vspace{-0.6cm}
\caption{\textbf{Per-scene qualitative results on neural radiance field reconstruction.} 
Despite using only 25 input views instead of the default 100, FLAIR reconstructs unseen views without artifacts and preserves fine details.}
\vspace{-0.3cm}
\label{fig:sup_NeRF}
\end{figure*}

In Fig.~\ref{tab:nerf_per_scene}, we visualize additional per-scene results for the novel view synthesis task. In the first row, FLAIR recovers fine-grained structures by selectively capturing the required frequency components and, through its frequency shifting mechanism, effectively reaching the desired high-frequency details. In the second, third rows, FLAIR maintains stable rendering quality even under a challenging setting where the model trains on only 25 views and renders 100 novel views. Under this hard regime, FLAIR suppresses noise, prevents structural leakage, and produces clean textures, whereas competing methods exhibit visible degradation. The qualitative observations are in line with the quantitative results. As shown in Table~\ref{fig:sup_NeRF}, FLAIR generally achieves consistent performance improvements across the per-scene evaluations with stable and reliable gains.

For SL2A~\cite{rezaeian2025sl2a}, the method adjusts polynomial degrees and low-rank linear layer ranks depending on the target task. However, SL2A does not provide NeRF-specific implementation details, and the official GitHub repository places the entry ``Add Novel View Synthesis (NeRF) code'' in the to-do list rather than providing an actual implementation. As of November~21, no operational NeRF implementation exists. We attempted several reasonable configurations, but training remained unstable, often resulting N/A outputs, so we do not report SL2A for the NeRF experiment. For a fair and comprehensive comparison, we also include the plug-and-play approach IGA~\cite{shiinductive}. Using IGA alone does not yield meaningful performance. Therefore, we combine IGA with a positional encoding module of comparable parameter size and report the resulting scores.
\begin{figure} \includegraphics[width=\linewidth,keepaspectratio]{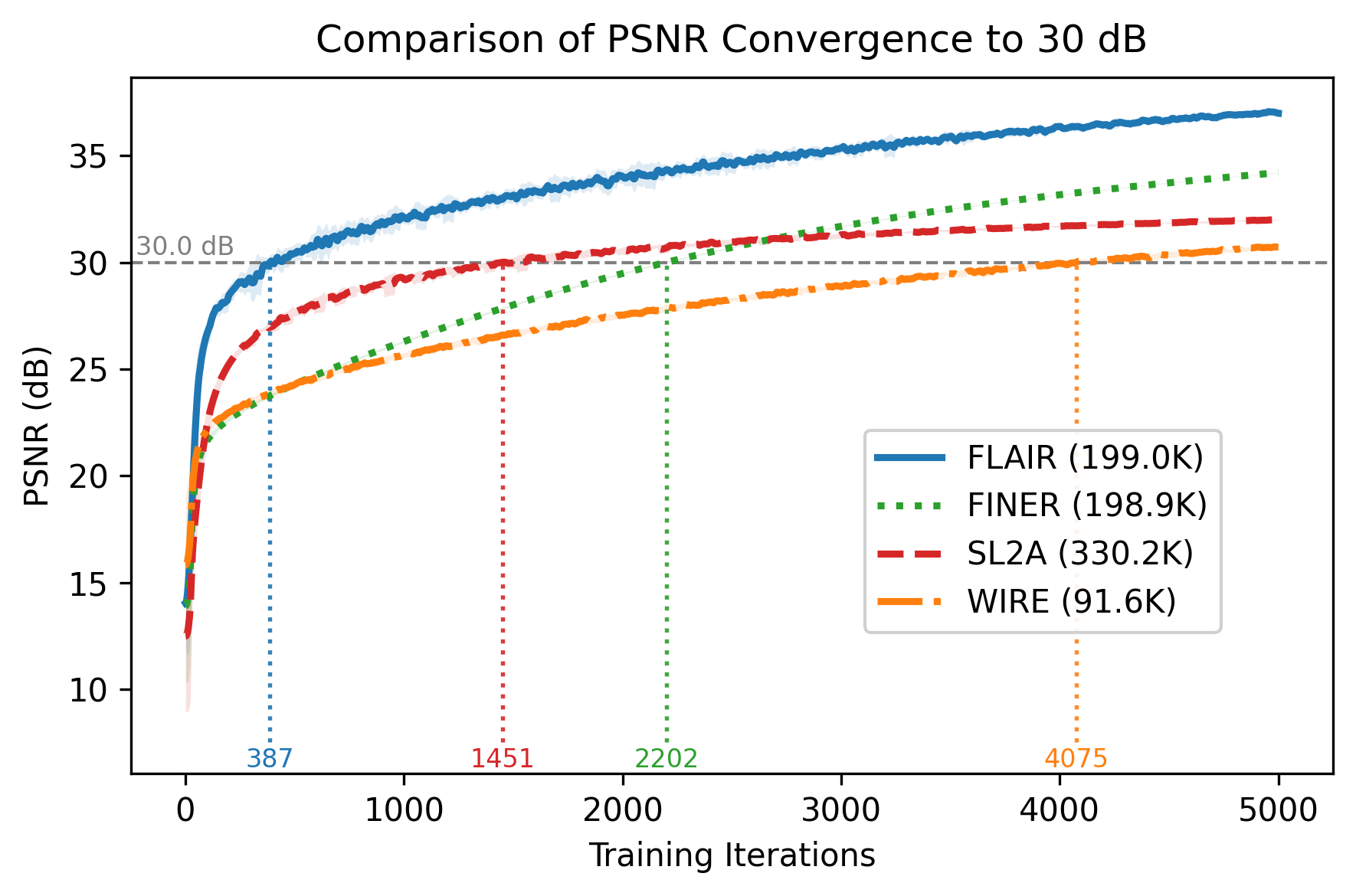}
\vspace{-0.7cm}\centering
\caption{\textbf{FLAIR achieves the fastest convergence to 30 dB.} FLAIR reaches the target quality with significantly fewer training iterations, indicating high training efficiency, and continues to improve to outperform all baselines in the final results. This advantage arises from the synergy between WEGE’s frequency-aware guidance and BLA’s adaptive representation, which allows the model to capture both low and high frequencies as needed.}
\label{fig:convergence} 
\end{figure}
\clearpage

\begin{table}
\caption{{\textbf{Quantitative comparison on the Tokyo dataset (fitting) and the LLFF dataset (novel view synthesis).}}
FLAIR (Small) achieves competitive performance with explicit methods while using significantly fewer parameters and a smaller model size. FLAIR (Large) achieves improved performance and attains the best results across both datasets.}
\vspace{-0.5cm}
\label{tab:Tokyo}
\begin{center}
\renewcommand{\arraystretch}{0.5}
\resizebox{\linewidth}{!}{
\begin{tabular}{c c c c c c c c}
    \toprule
    \multicolumn{8}{c}{\textbf{Tokyo Dataset} \; (Task: Fitting)} \\
    \midrule
     & Params~(M) $\downarrow$
     & Size~(MB)$\downarrow$ 
     & Steps~(k)
     & Time$\downarrow$ 
     & PSNR$\uparrow$ 
     & SSIM$\uparrow$
     & LPIPS$\downarrow$ \\
    \midrule
    Instant-NGP 
    & 3.67\,M
    & 10.54 
    & 30k
    & \best{3.4m} 
    & \second{28.38} 
    & 0.8240 
    & 0.208 \\
    
    NFFB 
    & \second{2.68}\,M
    & 10.21 
    & 30k
    & 30.1m 
    & 26.54 
    & 0.7675 
    & 0.234 \\
    
    Ours (Small)
    & \best{0.20}\,M
    & \best{1.51} 
    & 25k
    & \second{6.2m} 
    & 27.74 
    & \second{0.8554} 
    & \second{0.206} \\
    
    Ours (Large)
    & 3.25\,M
    & \second{10.19} 
    & 35k
    & 26.5m 
    & \best{34.70} 
    & \best{0.9525} 
    & \best{0.061} \\
    \midrule
    \multicolumn{8}{c}{\textbf{LLFF Dataset} \; (Task: NVS)} \\
    \midrule
    Instant-NGP 
    & 13.00\,M
    & 24.89 
    & 30k 
    & \best{6.4m} 
    & \second{30.20} 
    & \second{0.9524} 
    & \best{0.030} \\
    
    NFFB 
    & 3.89\,M
    & 47.11
    & 30k 
    & 42.4m 
    & 26.94
    & 0.8852 
    & 0.258 \\
    
    NeRF
    & \second{1.19}\,M
    & \second{4.55} 
    & 50k 
    & 35.1m  
    & 26.80 
    & 0.8794 
    & 0.256 \\
    
    Ours (Small)
    & \best{0.79}\,M
    & \best{3.03}  
    & 30k 
    & \second{14.0m} 
    & 28.24 
    & 0.8846 
    & 0.242 \\
    
    Ours (Large)
    & 2.55\,M
    & 9.74  
    & 37k 
    & 25.4m 
    & \best{31.14} 
    & \best{0.9566} 
    & \second{0.032} \\
    \bottomrule
\end{tabular}
}
\end{center}
\end{table}

\subsection{Extended Comparison with Explicit Methods}
\begin{figure*}
\centering
\includegraphics[width=\linewidth,keepaspectratio]{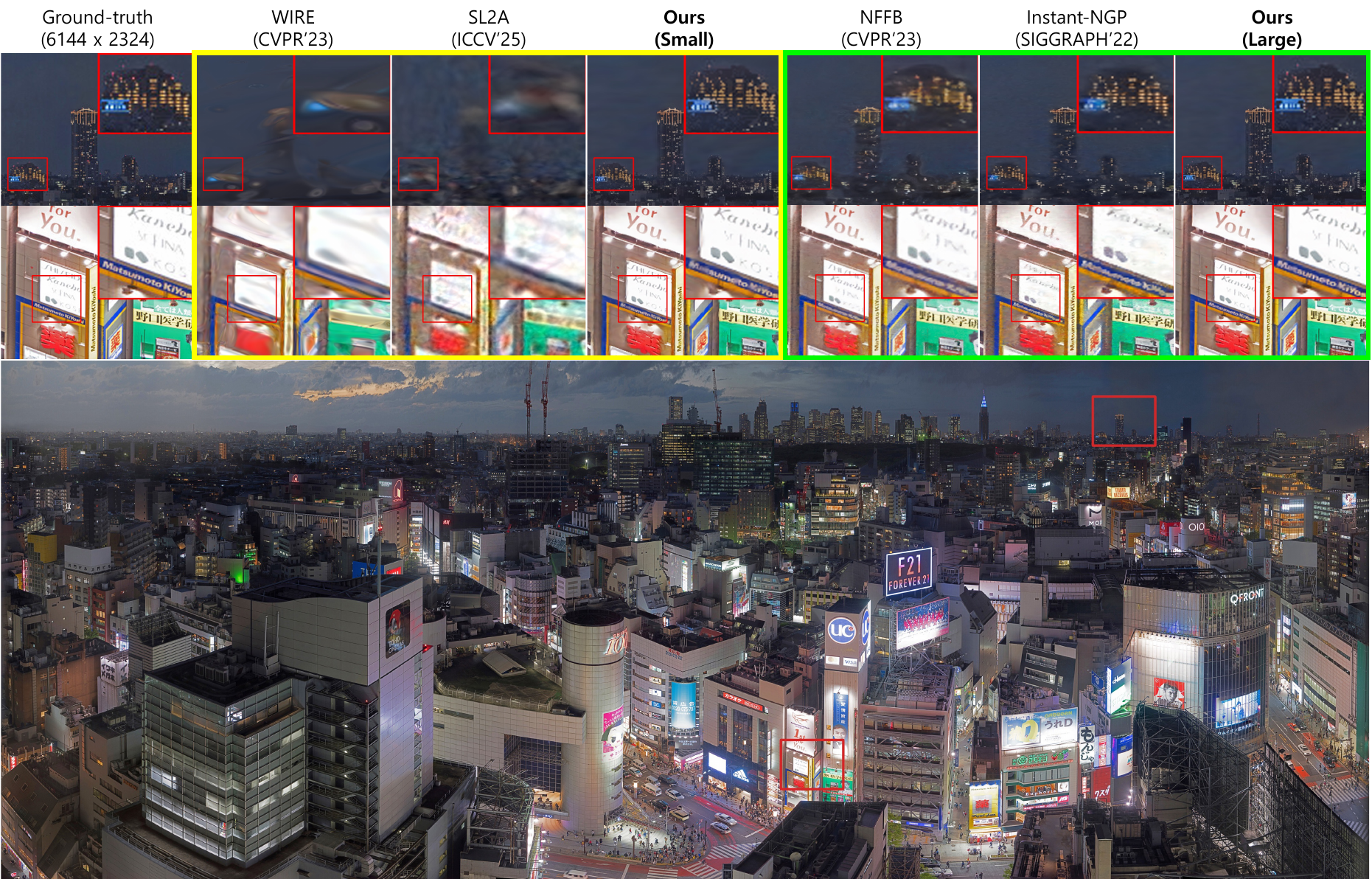}
\caption{\textbf{Visual comparison on a challenging gigapixel image (6144$\times$2324).}
FLAIR (Small) outperforms other implicit methods (yellow), and FLAIR (Large) produces more faithful results than explicit methods (green) at comparable computational cost.}
\label{fig:rebuttal1}
\end{figure*}

In Fig.~\ref{fig:fitting} of the main paper, the visual improvements over recent methods~\cite{rezaeian2025sl2a} appear limited due to saturation in the fitting task, where SSIM approaches 1.0. To better highlight the superiority of our model, we further evaluate on a more challenging high-resolution dataset, Tokyo~\cite{martel2021acorn}. As shown in Fig.~\ref{fig:rebuttal1}, the yellow box indicates configurations with matched layers and parameter budgets compared to prior INR methods, where our approach demonstrates strong performance. We further compare against explicit methods~\cite{muller2022instant,wu2023neural}, achieving competitive results with $10\times$ lower model capacity and computational cost, as summarized in Table~\ref{tab:Tokyo}.

\begin{figure*}
\centering
\includegraphics[width=\linewidth,keepaspectratio]
{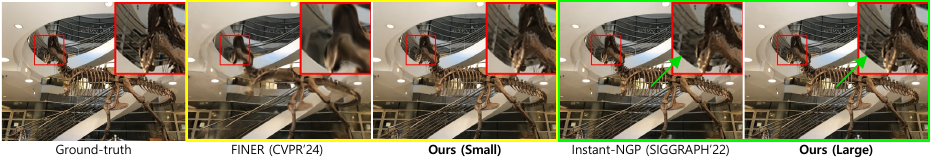}
\vspace{-0.6cm}
\caption{\textbf{Per-scene qualitative results on the LLFF dataset for neural radiance field reconstruction.} 
Beyond standard synthetic INR benchmarks, FLAIR reconstructs novel views while preserving fine details.}
\vspace{-0.3cm}
\label{fig:rebuttal2}
\end{figure*}

We further extend our evaluation on Neural Radiance Fields beyond standard synthetic benchmarks to the more realistic LLFF dataset~\cite{mildenhall2019local}, as shown in Fig.~\ref{fig:rebuttal2}. As illustrated in the figure, our method reconstructs finer details compared to both implicit and explicit methods (green arrows), demonstrating its broad applicability.

\subsection{Convergence Efficiency and Representation Compactness}
\noindent\textbf{Convergence Efficiency.}
Fig.~\ref{fig:convergence} evaluates how rapidly each method reaches a target quality of 30\,dB. FLAIR converges $4\times$ to $11\times$ faster than competing methods, and also achieves the best performance even at the final 5000 iterations. This demonstrates both its high training efficiency and its superior reconstruction accuracy.

\begin{figure} \includegraphics[width=\linewidth,keepaspectratio]{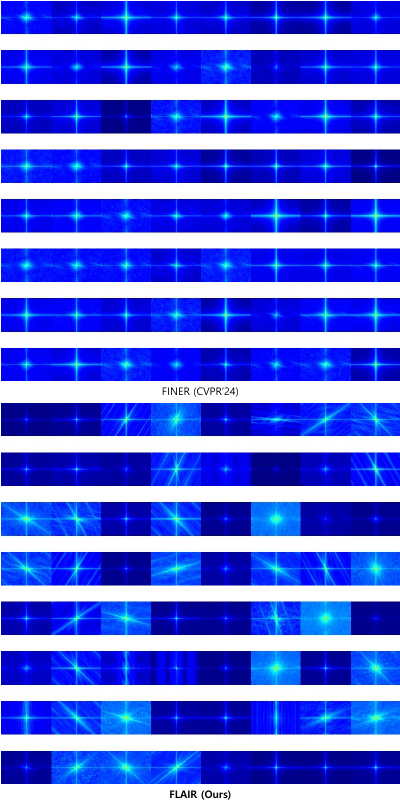}
\vspace{-0.7cm}\centering
\caption{\textbf{Fast fourier transform on learned neurons.}
Compared to the SOTA baseline FINER, whose frequency responses remain highly concentrated around the low-frequency center, FLAIR (ours) exhibits notably more diverse spectra. This indicates that FLAIR enables each neuron to learn the frequency components that are genuinely required.}
\label{fig:fft} 
\end{figure}

\begin{figure*} \includegraphics[width=\linewidth,keepaspectratio]{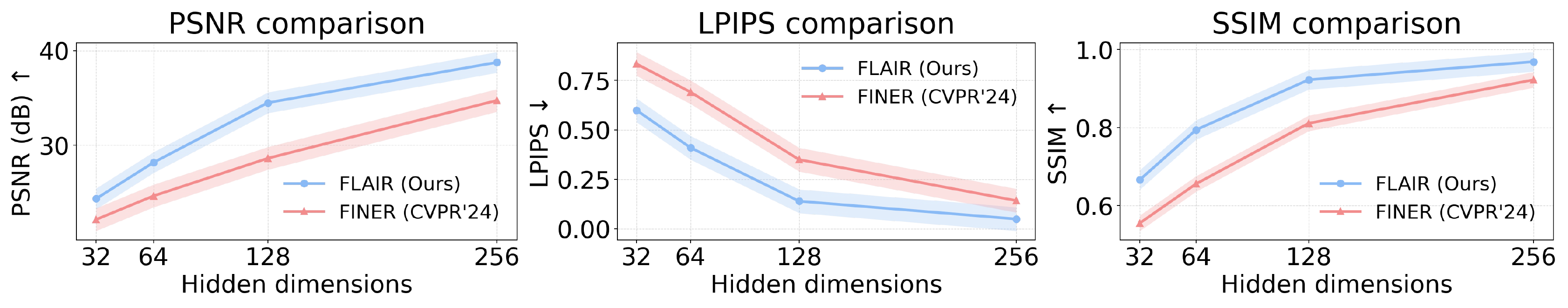}
\vspace{-0.3cm}\centering
\caption{\textbf{Comparison with the SOTA sinusoidal-based model FINER across multiple
hidden dimensions.} FLAIR achieves comparable performance to FINER while using
approximately half the hidden dimensionality for the same evaluation scores.
Since hidden dimension directly correlates with parameter count and FLOPs, this
demonstrates that FLAIR operates significantly better under constrained model budgets.}
\label{fig:sparse} 
\vspace{-0.5cm}
\end{figure*}

\noindent\textbf{Representation Compactness.}
FLAIR additionally exhibits strong representation compactness.
Fig.~\ref{fig:fft} visualizes the learned neurons via fast Fourier transform.
While the SOTA baseline FINER concentrates most neurons around similar low-frequency regions, revealing substantial redundancy, FLAIR learns diverse and distinct spectral patterns thanks to its band-limiting and frequency-shifting mechanisms. This diversity directly translates into robustness under reduced model capacity. As shown in Fig.~\ref{fig:sparse}, FINER degrades significantly when the hidden dimensionality is reduced, since its neurons do not cover sufficiently diverse frequency bands. In contrast, FLAIR continues to maintain comparable accuracy even when its model capacity is reduced to nearly half of that of FINER, as shown in Fig.~\ref{fig:sparse}. This indicates that FLAIR achieves a similar level of reconstruction quality with substantially fewer parameters and FLOPs.

\section{Applications and Limitations of FLAIR}
\label{sec:sup_E}

\subsection{Applications} 
\label{sec:sup_applications}
\begin{figure}[t]
\centering
\includegraphics[width=0.6\linewidth]{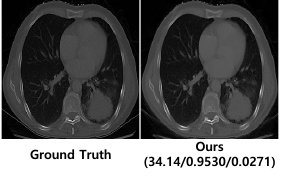}
\vspace{-0.6em}
\caption{\textbf{Computed tomography reconstruction.} FLAIR achieves high-quality reconstruction in sparse-view CT, effectively recovering continuous structures in medical imaging.}
\label{fig:medical}
\vspace{-1.2em}
\end{figure}

FLAIR exhibits broad applicability across diverse architectures and tasks. Rather than relying on complex designs, FLAIR achieves simultaneous localization and frequency selectivity using a single nonlinear activation, BLA. This allows BLA to integrate into arbitrary MLP architectures, such as NFFB and Scaffold-GS, and improves performance, as shown in Table~\ref{tab:integration}. Moreover, WEGE acts as a plug-and-play module compatible with various activation functions and enhances fine details, as illustrated in Fig.~\ref{fig:WEGE_fitting} and Fig.~\ref{fig:WEGE_SR}. By jointly using BLA and WEGE within an INR framework, FLAIR models continuous signals and is well-suited for domains involving continuous data, such as medical imaging (Fig.~\ref{fig:medical}).

\begin{table}[t]
\centering
\resizebox{0.9\linewidth}{!}{
\begin{tabular}{c c c c}
\hline
\multicolumn{4}{c}{\textbf{NFFB (CVPR'23)} \;|\; \textit{Tokyo dataset}} \\
\hline
Method & PSNR$\uparrow$ & SSIM$\uparrow$ & LPIPS$\downarrow$ \\
\hline
SIREN (Baseline) & 26.54 & 0.7675 & 0.234 \\
SIREN $\rightarrow$ BLA 
& 27.05 {\textcolor{red}{(+0.51)}} 
& 0.7824 {\textcolor{red}{(+0.0149)}} 
& 0.221 {\textcolor{red}{(-0.013)}} \\
\hline
\\[-0.8em]
\hline
\multicolumn{4}{c}{\textbf{Scaffold-GS (CVPR'24)} \;|\; \textit{Mip-NeRF360 bicycle}} \\
\hline
Method & PSNR$\uparrow$ & SSIM$\uparrow$ & LPIPS$\downarrow$ \\
\hline
ReLU (Baseline) & 24.44 & 0.6979 & 0.304 \\
ReLU $\rightarrow$ BLA 
& 24.78 {\textcolor{red}{(+0.34)}} 
& 0.7144 {\textcolor{red}{(+0.0165)}} 
& 0.285 {\textcolor{red}{(-0.019)}} \\
\hline
\end{tabular}
}
\vspace{-0.5em}
\caption{\textbf{BLA integrates into arbitrary MLP architectures.} Replacing SIREN in NFFB and ReLU in the two-layer MLP of Scaffold-GS with BLA consistently improves performance.}
\label{tab:integration}
\vspace{-1.2em}
\end{table}

\subsection{Limitations} 
\label{sec:sup_limitations}
\begin{figure} \includegraphics[width=\linewidth,keepaspectratio]{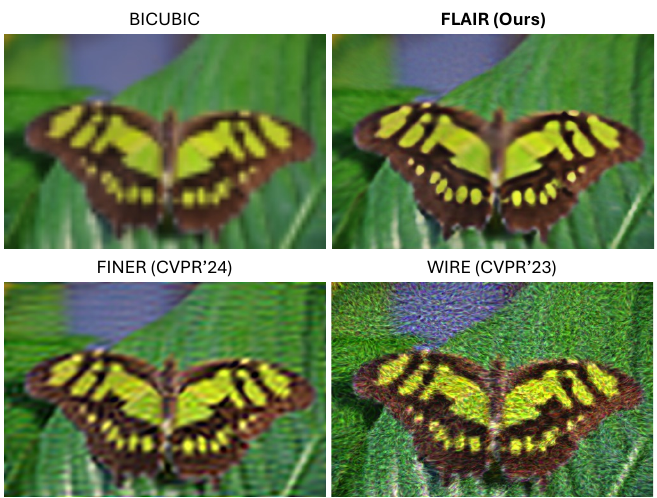}
\vspace{-0.7cm}\centering
\caption{\textbf{Super-Resolution under Extreme Degradation.}
Qualitative results on a challenging $\times8$ setting, showing that FLAIR suppresses noise more effectively than other methods  but still struggles under severe degradation.}
\label{fig:arb} 
\end{figure}

While INRs provide a compact and versatile representation that extends across diverse tasks, they also exhibit intrinsic limitations that FLAIR inherits. 
First, although FLAIR improves frequency selectivity and achieves strong quantitative results, as shown in Fig.~\ref{fig:arb}, it still struggles under extreme degradation scenarios in super-resolution. In such challenging regimes, diffusion-based approaches~\cite{jeong2025latent} demonstrate markedly higher stability, successfully reconstructing up to $\times 16$ magnification, where FLAIR leaves room for improvement. 
Second, despite being lightweight in structure, INRs are constrained to per-scene optimization, meaning that each new scene demands a full training procedure.
This stands in contrast to feed-forward architectures~\cite{oh2025sofono, chen2021learning}, which generalize across scenes and, once trained, require only a single forward pass for inference. Future work should focus on developing INR formulations that retain compactness while achieving stronger generalization and robustness across tasks and degradation levels.

\clearpage

\begin{table*}[t]
\centering
\caption{\textbf{Per-scene quantitative NeRF reconstruction results (8 scenes).}
\best{Red} and \second{blue} indicate best and second-best per column.}
\vspace{-0.2cm}
\label{tab:nerf_per_scene}

\resizebox{0.9\linewidth}{!}{%
\tiny
\setlength{\tabcolsep}{2pt}
\renewcommand{\arraystretch}{0.94}

\begin{tabular}{l|c|cccccccc|c}
\hline
\multirow{2}{*}{Methods} & \multirow{2}{*}{\#Params (K)} &
\multicolumn{8}{c|}{NeRF Scenes} & \multirow{2}{*}{Avg.} \\
\cline{3-10}
& & Drums & Chair & Ficus & Hotdog & Lego & Materials & Mic & Ship & \\
\hline
\multicolumn{11}{l}{\textit{PSNR $\uparrow$}} \\
\hline
IGA (ICML’25)      & 205.1 & 24.25 & 32.95 & 26.98 & \second{32.06} & 28.95 & 27.38 & 33.74 & \second{21.80} & 28.52 \\
FINER (CVPR’24)    & 198.9 & \second{24.91} & \second{33.38} & \second{27.28} & 32.71 & \second{29.77} & \second{27.64} & \second{34.42} & 21.63 & \second{28.97} \\
WIRE (CVPR’23)     &  91.6 & 23.87 & 30.68 & 25.63 & 31.15 & 27.58 & 25.50 & 32.23 & 20.95 & 27.20 \\
SIREN (NeurIPS’20) & 198.9 & 21.90 & 25.81 & 24.21 & 31.02 & 27.20 & 24.21 & 29.62 & 20.83 & 25.60 \\
\textbf{FLAIR (Ours)} & 199.0 & \best{25.03} & \best{34.03} & \best{27.33} & \best{33.12} & \best{30.00} & \best{28.29} & \best{34.75} & \best{21.93} & \best{29.31} \\
\hline
\multicolumn{11}{l}{\textit{SSIM $\uparrow$}} \\
\hline
IGA (ICML’25)      & 205.1 & 0.8858 & 0.9648 & 0.9417 & \second{0.9580} & 0.9349 & 0.9343 & 0.9771 & \second{0.7684} & 0.9217 \\
FINER (CVPR’24)    & 198.9 & \second{0.8995} & \second{0.9698} & \second{0.9466} & 0.9601 & \second{0.9479} & \second{0.9380} & \second{0.9796} & 0.7671 & \second{0.9261} \\
WIRE (CVPR’23)     &  91.6 & 0.8818 & 0.9482 & 0.9248 & 0.9410 & 0.9087 & 0.9047 & 0.9729 & 0.7249 & 0.9009 \\
SIREN (NeurIPS’20) & 198.9 & 0.8459 & 0.8955 & 0.9034 & 0.9475 & 0.9021 & 0.8967 & 0.9622 & 0.7320 & 0.8857 \\
\textbf{FLAIR (Ours)} & 199.0 & \best{0.9098} & \best{0.9752} & \best{0.9506} & \best{0.9666} & \best{0.9509} & \best{0.9520} & \best{0.9811} & \best{0.7734} & \best{0.9325} \\
\hline
\multicolumn{11}{l}{\textit{LPIPS $\downarrow$}} \\
\hline
IGA (ICML’25)      & 205.1 & 0.101 & 0.024 & 0.055 & \second{0.029} & 0.057 & 0.042 & 0.021 & 0.188 & 0.063 \\
FINER (CVPR’24)    & 198.9 & \second{0.055} & \second{0.017} & \second{0.037} & 0.032 & \best{0.027} & \second{0.033} & \best{0.012} & \second{0.139} & \second{0.044} \\
WIRE (CVPR’23)     &  91.6 & 0.071 & 0.031 & 0.059 & 0.047 & 0.049 & 0.066 & 0.020 & 0.178 & 0.065 \\
SIREN (NeurIPS’20) & 198.9 & 0.147 & 0.113 & 0.142 & 0.038 & 0.084 & 0.071 & 0.041 & 0.200 & 0.105 \\
\textbf{FLAIR (Ours)} & 199.0 & \best{0.053} & \best{0.015} & \best{0.036} & \best{0.022} & \second{0.033} & \best{0.026} & \second{0.013} & \best{0.126} & \best{0.041} \\
\hline
\end{tabular}
}
\vspace{-0.25cm}
\end{table*}

\clearpage

\begin{table*}[t]
\centering
\caption{\textbf{Per-scene quantitative image-fitting results on DIV2K (00–15).}
\best{Red} and \second{blue} indicate best and second-best values per column.}
\vspace{-0.2cm}
\label{tab:div2k}
\renewcommand{\arraystretch}{1.12}
\setlength{\tabcolsep}{3.5pt}

\resizebox{\linewidth}{!}{%
\begin{tabular}{l|c|cccccccccccccccc|c}
\hline
Methods & \#Params (K) &
00 & 01 & 02 & 03 & 04 & 05 & 06 & 07 & 08 & 09
& 10 & 11 & 12 & 13 & 14 & 15 & \textbf{Avg.} \\
\hline

\multicolumn{19}{l}{\textit{PSNR $\uparrow$}} \\
\hline
SL2A (ICCV’25)        & 330.2 &
\second{32.94} & \second{38.43} & \second{34.78} & \second{38.60} & \second{35.30} & \second{33.60} & \second{36.24} & \second{35.67} &
31.37 & \second{33.83} & \second{40.56} & \best{45.20} & 35.48 & \second{35.20} & \second{37.07} & \second{35.92} & \second{36.26} \\
IGA (ICML’25)         & 205.1 &
28.72 & 35.12 & 32.59 & 35.09 & 32.47 & 29.61 & 32.32 & 33.65 &
28.10 & 31.58 & 38.26 & 43.92 & 35.07 & 32.15 & 34.15 & 33.39 & 33.51 \\
FR (CVPR’24)          & 6299.9 &
23.13 & 26.64 & 25.09 & 28.38 & 24.64 & 23.84 & 25.67 & 25.99 &
22.69 & 23.84 & 31.17 & 32.75 & 25.79 & 26.24 & 27.67 & 25.96 & 26.22 \\
FINER (CVPR’24)       & 198.9 &
32.18 & 37.28 & 34.20 & 37.06 & 32.29 & 33.48 & 34.95 & 34.38 &
\second{32.03} & 33.08 & 37.32 & 42.03 & \second{36.30} & 33.94 & 36.31 & 33.35 & 35.01 \\
WIRE (CVPR’23)        & 91.6  &
28.15 & 33.83 & 30.81 & 35.62 & 30.28 & 29.01 & 31.41 & 31.37 &
28.21 & 28.74 & 36.90 & 40.57 & 34.24 & 31.25 & 33.89 & 30.40 & 32.17 \\
SIREN (NeurIPS’20)    & 198.9 &
31.15 & 36.74 & 33.50 & 37.44 & 33.28 & 31.84 & 34.18 & 34.31 &
30.24 & 31.56 & 36.83 & 41.39 & 36.17 & 32.59 & 35.43 & 33.09 & 34.36 \\
\textbf{FLAIR (Ours)} & 199.0 &
\best{34.75} & \best{38.88} & \best{37.19} & \best{40.45} & \best{38.12} & \best{35.09} & \best{37.64} & \best{37.34} &
\best{35.22} & \best{36.29} & \best{41.35} & \second{44.77} & \best{38.36} & \best{37.15} & \best{39.92} & \best{38.51} & \best{38.19} \\
\hline

\multicolumn{19}{l}{\textit{SSIM $\uparrow$}} \\   
\hline
SL2A (ICCV’25)        & 330.2 &
0.9271 & 0.9649 & 0.9361 & 0.9641 & 0.9480 & 0.9437 & 0.9555 & 0.9392 &
0.9319 & 0.9263 & 0.9747 & 0.9824 & 0.9345 & 0.9358 & 0.9613 & 0.9440 & 0.9481 \\
IGA (ICML’25)         & 205.1 &
\second{0.9563} & \second{0.9736} & \second{0.9708} & \second{0.9737} & \second{0.9796} & \second{0.9665} & \second{0.9668} & \second{0.9679} &
\second{0.9749} & \second{0.9732} & \best{0.9833} & \second{0.9839} & \best{0.9852} & \second{0.9746} & \second{0.9780} & \second{0.9786} & \second{0.9742} \\
FR (CVPR’24)          & 6299.9 &
0.8816 & 0.8999 & 0.9088 & 0.9325 & 0.8960 & 0.8915 & 0.8843 & 0.8996 &
0.9140 & 0.8801 & 0.9627 & 0.9050 & 0.9558 & 0.9121 & 0.9016 & 0.8628 & 0.9100 \\
FINER (CVPR’24)       & 198.9 &
0.9187 & 0.9542 & 0.9452 & 0.9564 & 0.9469 & 0.9442 & 0.9460 & 0.9385 &
0.9422 & 0.9331 & 0.9572 & 0.9747 & 0.9590 & 0.9330 & 0.9510 & 0.9302 & 0.9457 \\
WIRE (CVPR’23)        & 91.6  &
0.8446 & 0.9207 & 0.8831 & 0.9257 & 0.8825 & 0.8565 & 0.8612 & 0.8670 &
0.8647 & 0.8500 & 0.9226 & 0.9650 & 0.9277 & 0.8620 & 0.9015 & 0.8630 & 0.8874 \\
SIREN (NeurIPS’20)    & 198.9 &
0.9182 & 0.9577 & 0.9410 & 0.9584 & 0.9422 & 0.9221 & 0.9334 & 0.9384 &
0.9180 & 0.9249 & 0.9519 & 0.9714 & 0.9580 & 0.9151 & 0.9395 & 0.9310 & 0.9408 \\
\textbf{FLAIR (Ours)} & 199.0 &
\best{0.9663} & \best{0.9762} & \best{0.9737} & \best{0.9772} & \best{0.9809} & \best{0.9669} & \best{0.9730} & \best{0.9687} &
\best{0.9768} & \best{0.9739} & \second{0.9821} & \best{0.9850} & \second{0.9708} & \best{0.9753} & \best{0.9804} & \best{0.9798} & \best{0.9754} \\
\hline

\multicolumn{19}{l}{\textit{LPIPS $\downarrow$}} \\  
\hline
SL2A (ICCV’25) & 330.2 &
0.020 & \best{0.013} & 0.065 & 0.014 & 0.062 & 0.018 & \second{0.031} & 0.043 &
0.019 & 0.066 & 0.024 & \best{0.008} & 0.013 & 0.049 & 0.029 & 0.076 & 0.034 \\

IGA (ICML’25) & 205.1 &
0.036 & 0.019 & \best{0.019} & \second{0.013} & \best{0.014} & 0.038 & 0.042 & \best{0.011} &
0.041 & \second{0.018} & \second{0.017} & 0.015 & \second{0.005} & \second{0.024} & \best{0.016} & \second{0.026} & \second{0.022} \\

FR (CVPR’24) & 6299.9 &
0.176 & 0.103 & 0.145 & 0.046 & 0.168 & 0.156 & 0.170 & 0.141 &
0.184 & 0.122 & 0.038 & 0.064 & 0.033 & 0.111 & 0.103 & 0.226 & 0.124 \\

FINER (CVPR’24) & 198.9 &
\second{0.019} & 0.027 & 0.050 & 0.047 & 0.048 & \second{0.016} & 0.038 & 0.036 &
\second{0.012} & 0.047 & 0.070 & 0.032 & \second{0.005} & 0.046 & 0.045 & 0.085 & 0.039 \\

WIRE (CVPR’23) & 91.6 &
0.128 & 0.077 & 0.115 & 0.079 & 0.093 & 0.124 & 0.160 & 0.127 &
0.109 & 0.123 & 0.068 & 0.053 & 0.023 & 0.138 & 0.120 & 0.185 & 0.107 \\

SIREN (NeurIPS’20) & 198.9 &
0.034 & 0.031 & \second{0.047} & 0.047 &\second{0.039} & 0.032 & 0.077 & 0.051 &
0.042 & 0.044 & 0.083 & 0.041 & 0.013 & 0.082 & 0.067 & 0.070 & 0.050 \\

\textbf{FLAIR (Ours)} & 199.0 &
\best{0.013} & \second{0.016} & \best{0.019} & \best{0.012} & \best{0.014} & \best{0.014} & \best{0.028} & \second{0.026} &
\best{0.011} & \best{0.017} & \best{0.016} & \second{0.013} & \best{0.004} & \best{0.022} & \second{0.019} & \best{0.016} & \best{0.016} \\
\hline
\end{tabular}}
\end{table*}

\begin{table*}[t]
\centering
\caption{\textbf{Per-scene quantitative image-fitting results on Kodak (24 scenes).}
\best{Red} and \second{blue} indicate best and second-best values per column.}
\vspace{-0.2cm}
\label{tab:kodak}
\renewcommand{\arraystretch}{1.12}
\setlength{\tabcolsep}{3.5pt}

\resizebox{\linewidth}{!}{%
\begin{tabular}{l|c|cccccccccccccccccccccccc|c}
\hline
Methods & \#Params (K) &
1 & 2 & 3 & 4 & 5 & 6 & 7 & 8 & 9 & 10 & 11 & 12 & 13 & 14 & 15 & 16 & 17 & 18 & 19 & 20 & 21 & 22 & 23 & 24 & \textbf{Avg.} \\
\hline
\multicolumn{27}{l}{\textit{PSNR $\uparrow$}} \\
\hline
SL2A (ICCV’25) & 330.2 &
33.01 & \second{38.40} & \second{39.79} & \second{37.33} & 32.79 & \second{34.90} & \best{39.73} & 30.49 &
38.60 & \second{38.04} & \second{36.33} & \second{38.17} & 29.83 & 34.36 & 36.80 & \best{38.85} &
\best{39.01} & \best{34.36} & \second{35.91} & 36.42 & \second{35.67} & \second{35.01} & \best{40.73} & \second{32.74} & \second{36.14} \\
IGA (ICML’25) & 205.1 &
31.20 & 36.71 & 38.97 & 36.25 & 30.30 & 31.95 & 39.31 & 28.58 &
\best{38.80} & 37.88 & 34.03 & 37.17 & 27.29 & 31.95 & 35.28 & 36.31 &
36.83 & 31.53 & 34.32 & 35.75 & 33.54 & 33.92 & 39.23 & 30.73 & 34.49 \\
FR (CVPR’24) & 6299.9 &
24.68 & 29.57 & 30.69 & 29.25 & 24.01 & 26.23 & 29.40 & 22.88 &
29.59 & 28.85 & 27.24 & 30.37 & 22.42 & 26.09 & 30.03 & 28.73 &
29.57 & 25.78 & 27.04 & 29.61 & 27.05 & 27.53 & 31.81 & 24.51 & 27.62 \\
FINER (CVPR’24) & 198.9 &
\second{33.67} & 37.49 & 38.44 & 36.19 & \second{34.05} & 34.03 & 37.57 & \second{30.80} &
36.92 & 36.46 & 35.49 & 37.81 & \second{30.81} & \second{34.47} & \second{37.37} & 36.52 &
36.71 & \second{33.63} & 34.99 & \second{36.52} & 34.69 & 34.86 & 38.79 & 32.65 & 35.46 \\
WIRE (CVPR’23) & 91.6 &
26.30 & 31.22 & 33.04 & 30.74 & 26.69 & 28.00 & 30.43 & 23.28 &
31.64 & 30.24 & 29.20 & 30.25 & 25.80 & 27.03 & 30.63 & 30.49 &
29.23 & 25.45 & 27.33 & 28.83 & 28.45 & 29.37 & 30.83 & 26.60 & 28.80 \\
SIREN (NeurIPS’20) & 198.9 &
29.23 & 30.84 & 32.40 & 30.41 & 25.18 & 27.06 & 31.31 & 23.00 &
31.71 & 30.95 & 28.05 & 31.53 & 22.71 & 26.98 & 30.49 & 29.28 &
30.41 & 26.03 & 27.20 & 29.73 & 27.22 & 28.59 & 32.47 & 24.54 & 28.64 \\
\textbf{FLAIR (Ours)} & 199.0 &
\best{36.11} & \best{40.34} & \best{40.93} & \best{39.89} & \best{35.01} & \best{36.20} & \second{39.32} & \best{32.89} &
\second{38.63} & \best{38.10} & \best{36.75} & \best{39.10} & \best{33.67} & \best{35.85} & \best{38.06} & \second{38.76} &
\second{37.29} & \second{33.63} & \best{35.95} & \best{37.22} & \best{36.32} & \best{36.45} & \second{39.90} & \best{34.71} & \best{37.12} \\
\hline

\multicolumn{27}{l}{\textit{SSIM $\uparrow$}} \\
\hline
SL2A (ICCV’25) & 330.2 &
0.9255 & 0.9320 & 0.9558 & 0.9307 & 0.9260 & 0.9298 & 0.9658 & 0.8980 &
0.9430 & 0.9399 & 0.9330 & 0.9382 & 0.8828 & 0.9264 & 0.9222 & 0.9579 &
0.9563 & 0.9177 & 0.9278 & 0.9207 & 0.9305 & 0.9145 & 0.9576 & 0.8985 & 0.9304 \\
IGA (ICML’25) & 205.1 &
\second{0.9631} & \second{0.9474} & \second{0.9754} & \second{0.9690} & \second{0.9686} & \second{0.9321} & \second{0.9743} & \second{0.9501} &
\second{0.9646} & \best{0.9781} & \second{0.9461} & \second{0.9632} & \second{0.9336} & \second{0.9599} & \best{0.9828} & \second{0.9636} &
\best{0.9722} & \second{0.9311} & \second{0.9424} & \second{0.9822} & \second{0.9630} & \second{0.9459} & 0.9630 & \second{0.9566} & \second{0.9596} \\
FR (CVPR’24) & 6299.9 &
0.8404 & 0.8029 & 0.9478 & 0.8822 & 0.8778 & 0.8008 & 0.9514 & 0.9031 &
0.8945 & 0.8981 & 0.8373 & 0.8911 & 0.7742 & 0.8843 & 0.9509 & 0.8299 &
0.9285 & 0.8369 & 0.8450 & 0.9120 & 0.8800 & 0.8529 & \best{0.9678} & 0.8855 & 0.8781 \\
FINER (CVPR’24) & 198.9 &
0.9188 & 0.9155 & 0.9345 & 0.9102 & 0.9413 & 0.9135 & 0.9590 & 0.9157 &
0.9195 & 0.9257 & 0.9150 & 0.9294 & 0.9173 & 0.9213 & 0.9282 & 0.9271 &
0.9384 & 0.9207 & 0.9129 & 0.9254 & 0.9239 & 0.9161 & 0.9434 & 0.9185 & 0.9246 \\
WIRE (CVPR’23) & 91.6 &
0.6672 & 0.7730 & 0.8742 & 0.7921 & 0.8121 & 0.7823 & 0.8654 & 0.7013 &
0.8432 & 0.8310 & 0.8017 & 0.8194 & 0.7542 & 0.7308 & 0.8222 & 0.7892 &
0.8212 & 0.6992 & 0.7668 & 0.8188 & 0.8275 & 0.7680 & 0.8669 & 0.8001 & 0.7928 \\
SIREN (NeurIPS’20) & 198.9 &
0.8454 & 0.7627 & 0.8591 & 0.7829 & 0.7362 & 0.7551 & 0.8975 & 0.6887 &
0.8669 & 0.8420 & 0.7424 & 0.8206 & 0.5735 & 0.7114 & 0.8084 & 0.7489 &
0.8479 & 0.7032 & 0.7405 & 0.8336 & 0.8035 & 0.7300 & 0.8967 & 0.7027 & 0.7792 \\
\textbf{FLAIR (Ours)} & 199.0 &
\best{0.9795} & \best{0.9695} & \best{0.9785} & \best{0.9693} & \best{0.9703} & \best{0.9667} & \best{0.9792} & \best{0.9591} &
\best{0.9770} & \second{0.9605} & \best{0.9556} & \best{0.9786} & \best{0.9559} & \best{0.9601} & \second{0.9569} & \best{0.9689} &
\second{0.9593} & \best{0.9314} & \best{0.9428} & \best{0.9909} & \best{0.9645} & \best{0.9471} & \second{0.9631} & \best{0.9597} & \best{0.9644} \\
\hline

\multicolumn{27}{l}{\textit{LPIPS $\downarrow$}} \\
\hline
SL2A (ICCV’25) & 330.2 &
0.056 & 0.047 & 0.037 & 0.071 & 0.039 & 0.064 & \best{0.022} & 0.064 &
\second{0.051} & \best{0.046} & \best{0.062} & 0.068 & 0.108 & \best{0.054} & \second{0.063} & \second{0.050} &
\best{0.030} & \best{0.052} & \best{0.083} & 0.080 & 0.080 & 0.072 & \second{0.045} & 0.092 & 0.060 \\
IGA (ICML’25) & 205.1 &
\second{0.045} & 0.043 & \second{0.027} & \second{0.055} & 0.033 & \second{0.059} & \second{0.030} & 0.051 &
\best{0.018} & 0.069 & 0.054 & \second{0.050} & 0.099 & 0.086 & 0.066 & \second{0.050} &
0.082 & 0.086 & 0.114 & \best{0.035} & \second{0.066} & \best{0.032} & 0.050 & \second{0.064} & \second{0.056} \\
FR (CVPR’24) & 6299.9 &
0.206 & 0.201 & 0.087 & 0.111 & 0.202 & 0.186 & 0.092 & 0.170 &
0.062 & 0.106 & 0.208 & 0.086 & 0.385 & 0.165 & 0.085 & 0.183 &
0.096 & 0.175 & 0.127 & 0.104 & 0.157 & 0.200 & 0.051 & 0.213 & 0.151 \\
FINER (CVPR’24) & 198.9 &
0.051 & \second{0.039} & 0.032 & 0.073 & \second{0.029} & 0.060 & 0.036 & \second{0.045} &
0.076 & \second{0.053} & 0.081 & 0.061 & \second{0.086} & 0.064 & 0.088 & 0.060 &
0.074 & \second{0.072} & 0.104 & 0.081 & 0.083 & 0.072 & 0.084 & 0.078 & 0.065 \\
WIRE (CVPR’23) & 91.6 &
0.445 & 0.383 & 0.182 & 0.332 & 0.224 & 0.289 & 0.202 & 0.383 &
0.298 & 0.331 & 0.350 & 0.379 & 0.373 & 0.442 & 0.288 & 0.362 &
0.302 & 0.440 & 0.467 & 0.272 & 0.413 & 0.321 & 0.223 & 0.280 & 0.333 \\
SIREN (NeurIPS’20) & 198.9 &
0.167 & 0.423 & 0.230 & 0.360 & 0.309 & 0.353 & 0.188 & 0.441 &
0.290 & 0.324 & 0.432 & 0.386 & 0.618 & 0.475 & 0.332 & 0.456 &
0.280 & 0.436 & 0.500 & 0.252 & 0.446 & 0.446 & 0.222 & 0.491 & 0.369 \\
\textbf{FLAIR (Ours)} & 199.0 &
\best{0.018} & \best{0.021} & \best{0.024} & \best{0.052} & \best{0.019} & \best{0.052} & \best{0.022} & \best{0.034} &
\second{0.051} & 0.062 & \second{0.065} & \best{0.046} & \best{0.071} & \second{0.059} & \best{0.062} & \best{0.048} &
\second{0.072} & 0.098 & \second{0.101} & \second{0.061} & \best{0.063} & \second{0.070} & \best{0.043} & \best{0.061} & \best{0.054} \\
\hline
\end{tabular}}
\vspace{-0.25cm}
\end{table*}

\clearpage


\end{document}